\documentclass[final]{cvpr}

\usepackage{times}
\usepackage{epsfig}
\usepackage{graphicx}
\usepackage{amsmath}
\usepackage{amssymb}

\usepackage[utf8]{inputenc} %
\usepackage[T1]{fontenc}    %
\usepackage{url}            %
\usepackage{booktabs}       %
\usepackage{amsfonts}       %
\usepackage{nicefrac}       %
\usepackage{microtype}      %
\usepackage{graphicx}
\usepackage{subcaption}
\usepackage{placeins}
\usepackage[dvipsnames]{xcolor}

\usepackage{float}

\usepackage{multirow, makecell}
\usepackage{numprint}
\npthousandsep{,}
\usepackage{nicefrac}
\usepackage{capt-of}

\usepackage[pagebackref=true,breaklinks=true,colorlinks,bookmarks=false]{hyperref}

\def\cvprPaperID{00283} %
\def\confYear{CVPR 2021}

\newlength{\tabcolsepdefault}
\DeclareMathOperator{\vectorspan}{span}
\DeclareMathOperator{\card}{card}

\newcommand\customparagraph[1]{\vspace{0.4em}\noindent\textbf{#1}}

\begin{document}
	
\title{Privacy-Preserving Image Features via Adversarial Affine Subspace Embeddings}

\author{%
	Mihai Dusmanu$^1$ \hspace{2mm} Johannes L. Sch\"onberger$^2$ \hspace{2mm} Sudipta N. Sinha$^2$ \hspace{2mm} Marc Pollefeys$^{1,2}$\\
	$^1$ Department of Computer Science, ETH Z\"urich \hspace{5mm} $^2$ Microsoft \\
}

\maketitle

\begin{abstract}
Many computer vision systems require users to upload image features to the cloud for processing and storage.
These features can be exploited to recover sensitive information about the scene or subjects, e.g., by reconstructing the appearance of the original image.
To address this privacy concern, we propose a new privacy-preserving feature representation.
The core idea of our work is to drop constraints from each feature descriptor by embedding it within an affine subspace containing the original feature as well as adversarial feature samples.
Feature matching on the privacy-preserving representation is enabled based on the notion of subspace-to-subspace distance.
We experimentally demonstrate the effectiveness of our method and its high practical relevance for the applications of visual localization and mapping as well as face authentication.
Compared to the original features, our approach makes it significantly more difficult for an adversary to recover private information.
\end{abstract}
\vspace{-5pt}
\section{Introduction}
\label{sec:introduction}

Image feature extraction and matching are two fundamental steps in many computer vision applications, such as 3D reconstruction~\cite{Lowe2004Distinctive,Agarwal2011Building}, image retrieval \cite{Lowe2004Distinctive,Sivic2003Video}, or face recognition~\cite{Turk1991Eigenfaces}.
Image features can be categorized into low-level~\cite{Lowe2004Distinctive,Dalal2005Histograms}, mid-level~\cite{Sivic2003Video,Boureau2010Learning} or high-level~\cite{Sharif2014CNN,Zeiler2014Visualizing} depending on their information content and receptive field.
Furthermore, features can be hand-crafted or learned using data-driven techniques.
However, they are almost always represented as vectors in high-dimensional feature spaces.
Multiple feature vectors are then compared using appropriate distance metrics, which forms the basis of nearest neighbor search or other retrieval and recognition techniques.

\begin{figure}[t]
	\centering
	\includegraphics[width=\linewidth]{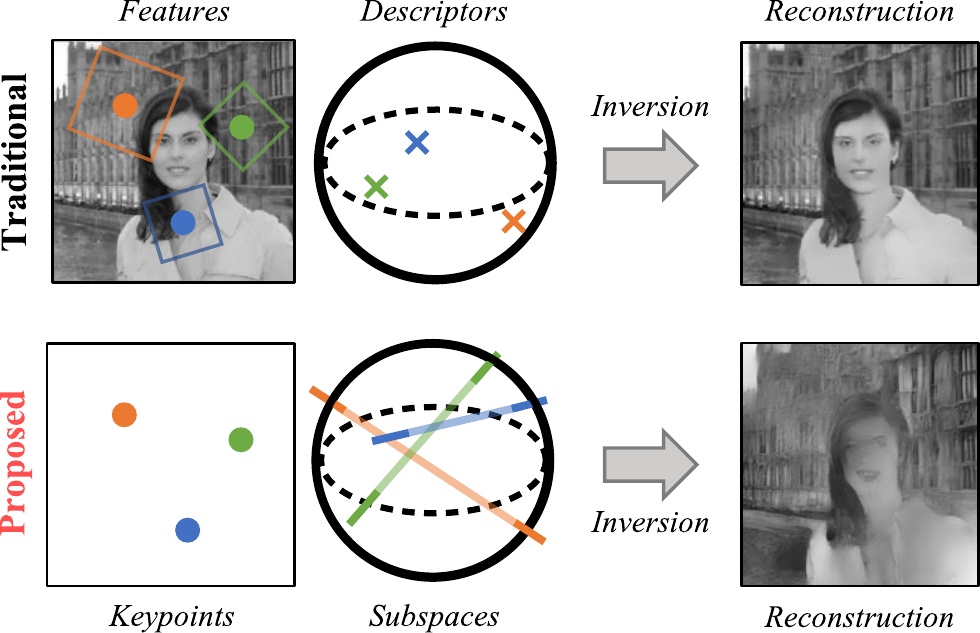}
	\caption{
		{\bf Privacy-Preserving Image Features.} Inversion of traditional local image features is a privacy concern in many applications. Our proposed approach obfuscates the appearance of the original image by lifting the descriptors to affine subspaces. Distance between the privacy-preserving subspaces enables efficient matching of features. The same concept can be applied to other domains such as face features for biometric authentication. Image credit: \textit{laylamoran4battersea} (Layla Moran).}
	\label{fig:teaser}
	\vspace{-5pt}
\end{figure}

Recently, there has been rapid progress in \textit{feature inversion} methods that reconstruct the image appearance from features extracted in the original image~\cite{Weinzaepfel2011Reconstructing,Dosovitskiy2016Inverting,Zhmoginov2016Inverting,Mai2018Reconstruction} as shown in Figure~\ref{fig:teaser}.
This raises serious privacy concerns, since images may contain sensitive information about the scene or subjects.
Increased awareness of these privacy issues has spurred significant efforts to develop privacy-preserving machine learning systems.
In recent years, researchers have proposed a large body of approaches to tackle the various aspects of the problem, including homomorphic cryptosystems~\cite{Paillier1999Public}, differential privacy~\cite{Dwork2006Calibrating}, federated learning~\cite{Kairouz2019Advances}, and specific solutions for camera localization~\cite{Speciale2019a,Speciale2019b}.

In this paper, we propose a new feature representation suitable for visual recognition and matching tasks that makes it significantly more difficult for an adversary to reconstruct the image contents.
Our approach has only marginal computational overhead, which makes it amenable to a wide range of practical scenarios.
The core idea behind our method is to represent a descriptor point in $\mathbb{R}^n$ as an affine subspace of $\mathbb{R}^n$ passing through the original point.
We refer to this process as \textit{lifting}.
The chosen dimension of the subspace determines a trade-off between accuracy, runtime, and the level of privacy of the feature representation.
To make inverting the representation difficult, we propose a strategy for constructing a lifted subspace containing additional adversarial feature points.
We empirically demonstrate strong privacy preservation even for low-dimensional affine subspaces.
Pairwise feature comparison is a fundamental step required in many recognition tasks. In our proposed framework, such comparisons are done directly on the lifted subspaces based on either point-to-subspace or subspace-to-subspace distance.

The paper is organized as follows.
First, we formally present the idea of lifting and the technique for matching lifted features.
Next, we analyze the performance of these features for two applications: a) image matching for visual localization and mapping as well as b) face authentication.
Finally, we demonstrate that our proposed representation is resilient to potential privacy attacks.
The code of our method and the evaluation protocol will be released as open-source.

\section{Related Work}
\label{sec:related}

We first review image features used for applications considered in this paper.
We then discuss existing work about privacy attacks on image features and defense mechanisms.

\customparagraph{Feature Descriptors.}
In the traditional local feature extraction paradigm, after keypoint detection and shape estimation, normalized image patches are extracted from images. Feature description takes a patch as input and outputs an $n$-dimensional vector.
Handcrafted local descriptors are based on direct pixel sampling~\cite{Calonder2011BRIEF} or a histogram of image gradients~\cite{Lowe2004Distinctive, Dalal2005Histograms}.
Recent advances in deep learning have led to descriptors based on convolutional neural networks (CNNs). Such learnable descriptors are trained using triplet~\cite{Balntas2016Learning} or list-wise~\cite{He2018Local} losses and hard-negative mining techniques~\cite{Mishchuk2017Working}.
Local features have been successfully used for tasks such as large-scale 3D reconstruction from crowd-sourced images~\cite{Agarwal2011Building} and image retrieval~\cite{Jegou2010Aggregating,schoenberger2015detail}.

Face recognition methods start by face detection and alignment to obtain a canonical face image~\cite{Zhang2016Joint}.
Subsequently, a well chosen low-dimensional subspace of pixel-space can provide good recognition performance~\cite{Turk1991Eigenfaces}.
More recently, CNN-based features have become the de facto choice for face descriptors.
These networks are trained using different classification losses~\cite{Taigman2014DeepFace,Liu2017SphereFace,Deng2018ArcFace}.

\customparagraph{Feature Subspaces.}
Wang~\etal~\cite{Wang2014Affine} also use a subspace representation for feature matching.
Different to their method, we consider \emph{affine} instead of linear subspaces.
Accordingly, our distance function is not based on principal angles but on the closest pair of points between the two subspaces.
Further, contrary to grouping descriptors of similar patches together to improve matching performance, we add adversarial descriptors to the subspaces to improve privacy.

\customparagraph{Feature Inversion and Compromising Privacy.} Weinzaepfel~\etal~\cite{Weinzaepfel2011Reconstructing} proposed a method for reconstructing images from local image features using a database of patches with associated descriptors.
Dosovitsky and Brox~\cite{Dosovitskiy2016Inverting,Dosovitskiy2016Generating} extended on this work by using a CNN and perceptual losses, while Pittaluga~\etal~\cite{Pittaluga2019} showed that it was possible to recover detailed images from sparse 3D point clouds reconstructed using structure-from-motion.
Similarly, Zhmoginov and Sandler~\cite{Zhmoginov2016Inverting} and Mai~\etal~\cite{Mai2018Reconstruction} proposed methods for reconstructing face images from their descriptors.
Moreover, they showed that the reconstructed images could even be used by an attacker to fool an authentication system.

\customparagraph{Privacy-Preserving Methods.} \textit{Differential privacy}~\cite{Dwork2006Calibrating} expands upon Dalenius~\cite{Dalenius1977Towards} by formalizing the problem of querying a database without inadvertently releasing information distinguishing the individual entries in the database.
An extended overview can be found in~\cite{Dwork2008Differential}.
Instead of protecting information leakage from a database, our scenario is quite different in that we are interested in protecting private information in the query as well as contributing new information to a database in a privacy-preserving manner.

McMahan~\etal~\cite{McMahan2016Communication} introduced \textit{federated learning}, a distributed client-server framework for training a model, where
training data remains with the clients, thus offering better privacy guarantees.
Kairouz~\etal~\cite{Kairouz2019Advances} reviews the topic and discusses open problems.
In contrast, we address a different setting, where tasks require image features computed by clients to be shared with the server.
In this context, our approach makes it difficult to recover private image information from the shared features.

Existing works on local features process images encrypted using different homomorphic cryptosystems~\cite{Qin2014Towards,Hsu2012Image} in the cloud.
Jiang~\etal~\cite{Jiang2017Secure} proposed an alternative by additively splitting the image into two ciphertext matrices using a private prime modulus.
These methods guarantee that the original images remain private, but they do not prevent information leakage by inverting the obtained local features.
One could also use $\ell_2$ distance computation on encrypted feature vectors~\cite{Kim2020Efficient,boddeti2018secure,engelsma2020hers}.
However, recent works regarding homomorphic representation search~\cite{boddeti2018secure,engelsma2020hers} remain computationally expensive, while our method only comes with marginal overhead.
Furthermore, these cryptosystems provide security through encryption, where a breach of the secret keys is a privacy risk.
In contrast, our system does not have the same single point of failure and provides parameters to trade off accuracy, runtime, and privacy.

Speciale~\etal~\cite{Speciale2019a,Speciale2019b} proposed solutions tailored to image-based localization, where geometric information is concealed by lifting 2D or 3D points to randomly oriented lines passing through the original locations.
Recent work extends on the same idea to solve the full structure-from-motion problem~\cite{geppert2020privacy,shibuya2020privacy}.
We draw inspiration from their approach, but instead lift feature descriptors to higher dimensional affine subspaces to conceal appearance information.

\section{Method}
\label{sec:method}

In this paper, we will represent features from a particular domain as vectors in $\mathbb{R}^n$, where $n$ is the dimensionality of the original feature space.
We denote $\vectorspan (v_1, \dots, v_m) = \{ \sum_{i=1}^m \lambda_i v_i | \lambda_i \in \mathbb{R} \}$ the linear span of a set of vectors $v_i \in \mathbb{R}^n$.
An $m$-dimensional affine subspace $\mathcal{A}$ will be represented as the vector sum of a translation vector $a_0$ and a linear subspace $\vectorspan (a_1, \dots, a_m)$, giving $\mathcal{A} = a_0 + \vectorspan (a_1, \dots, a_m)$.
The core idea of our method is to lift the original feature vector or descriptor $d \in \mathbb{R}^n$ to an $m$-dimensional affine subspace $\mathcal{D} \subset \mathbb{R}^n$ satisfying $d \in \mathcal{D}$.
We denote the lifted affine subspace representation as private features.
There are two major requirements that we must address.
Firstly, we need to define a distance function that can be used to reliably and efficiently compare two features in this new representation.
Secondly, we must construct the affine subspace in a way that effectively conceals the original feature vector $d$ and makes it difficult for an
attacker to carry out a successful privacy attack aiming to recover the vector $d$ given the private representation $\mathcal{D}$.

\subsection{Distance Functions}

Most applications require feature descriptor comparison, which is accomplished using appropriate pairwise distance measures. In our analysis, we restrict ourselves to the Euclidean distance (denoted $\lVert \cdot \rVert$) as it is most commonly used in practice. To compute the distance between private features, we either use the point-to-subspace or subspace-to-subspace distance. Note that both distances are upper bound by the original point-to-point distance.

\customparagraph{Point-to-Subspace Distance.} To compute the distance between a private descriptor $d$ represented as an affine subspace $\mathcal{D}$ and a regular descriptor $e$, one can use the point-to-subspace distance defined as:
\begin{equation}
	\text{dist}(\mathcal{D}, e) = \min_{x \in \mathcal{D}} \lVert e - x \rVert = \lVert e - p_\perp^\mathcal{D} (e) \lVert \enspace ,
\end{equation}
where $p_\perp^\mathcal{D}(e)$ denotes the orthogonal projection of $e$ onto $\mathcal{D}$.

\customparagraph{Subspace-to-Subspace Distance.} To compute the distance between two private descriptors $d, e$ represented as affine subspaces $\mathcal{D}, \mathcal{E}$ of dimensions $m_\mathcal{D}, m_\mathcal{E}$, one can use the subspace-to-subspace distance defined as:
\begin{equation}
\text{dist}(\mathcal{D}, \mathcal{E}) = \min_{x \in \mathcal{D}, y \in \mathcal{E}} \lVert y - x \rVert \enspace .
\end{equation}
Let us denote a closest pair of points in the two subspaces as $x^* \in \mathcal{D}$ and $y^* \in \mathcal{E}$, respectively. Then, we have:
\begin{equation}
x^* = d_0 + \sum_{i=1}^{m_\mathcal{D}} \alpha_i d_i \enspace ,
y^* = e_0 + \sum_{i=1}^{m_\mathcal{E}} \beta_i e_i \enspace ,
\label{eq:min-xy}
\end{equation}
where $\boldsymbol\alpha \in \mathbb{R}^{m_\mathcal{D}}, \boldsymbol\beta \in \mathbb{R}^{m_\mathcal{E}}$.
In the following derivation, we assume that both subspaces have the same dimension $m = m_\mathcal{D} = m_\mathcal{E}$ for simplicity.
A sufficient and necessary condition for $\text{dist}(\mathcal{D}, \mathcal{E}) = \lVert y^* - x^* \rVert$ is that the line $y^* - x^*$ is orthogonal to both $\mathcal{D}$ and $\mathcal{E}$:
\begin{equation}
\begin{cases}
	(y^* - x^*)^T d_i = 0 \\
	(y^* - x^*)^T e_i = 0
\end{cases} \enspace ,
\end{equation}
which can be rewritten as:
\begin{equation}
\begin{cases}
	(e_0 - d_0)^T d_i = \sum_{j = 1}^m \alpha_j d_i^T d_j + \sum_{j = 1}^m \beta_j (-d_i^T e_j)  \\
	(e_0 - d_0)^T e_i = \sum_{j = 1}^m \alpha_j e_i^T d_j + \sum_{j = 1}^m \beta_j (-e_i^T e_j) \enspace .
\end{cases}
\end{equation}
This system can be formulated in a more compact form:
\begin{equation}
	\begin{bmatrix}
	D D^T & -D E^T \\
	E D^T & -E E^T
	\end{bmatrix} \begin{bmatrix}
	\boldsymbol\alpha \\
	\boldsymbol\beta
	\end{bmatrix} = \begin{bmatrix}
	D \\
	E
	\end{bmatrix} (e_0 - d_0) \enspace ,
\end{equation}
where $D = \begin{bmatrix} d_1 \dots d_m\end{bmatrix}^T, E = \begin{bmatrix} e_1 \dots e_m\end{bmatrix}^T \in M_{m \times n}(\mathbb{R})$.

If the bases of the subspaces are orthonormal ($D D^T = E E^T = I$), the system further simplifies to:
\begin{equation}
\begin{bmatrix}
I & -D E^T \\
E D^T & -I
\end{bmatrix} \begin{bmatrix}
\boldsymbol\alpha \\
\boldsymbol\beta
\end{bmatrix} = \begin{bmatrix}
D \\
E
\end{bmatrix} (e_0 - d_0) \enspace . \label{eq:system}
\end{equation}

Thus, finding the subspace-to-subspace distance requires solving a linear system with $2m$ unknowns and equations. Let $M = -D E^T$. Under the assumption that the matrix $N = I - M M^T$ is invertible, the block-matrix inversion formula can be used to rewrite Eq.~\ref{eq:system} as follows:
\begin{align}
\begin{bmatrix}
\boldsymbol\alpha \\
\boldsymbol\beta
\end{bmatrix} =
\begin{bmatrix}
N^{-1} & N^{-1} M \\
-M^T N^{-1} & -M^T N^{-1} M - I
\end{bmatrix} \begin{bmatrix}
D \\
E
\end{bmatrix} (e_0 - d_0) \enspace .
\end{align}
The solutions to $\boldsymbol\alpha, \boldsymbol\beta$ can be substituted into Eq.~\ref{eq:min-xy} to obtain the subspace-to-subspace distance as $\lVert x^* - y^* \rVert$.
Note that the problem can also be formulated using the dual representation of a subspace as the intersection of $n - m$ hyperplanes.
We provide a derivation of the dual formulation and further discussion in the supplementary material.

\subsection{Affine Subspace Embedding}

Each subspace embedding is defined by a translation vector $d_0$ and a basis $\{d_1, \dots, d_m\}$.
The choice of these and the distribution of the original descriptors has significant impact on the effectiveness of our approach and the required dimensionality $m$ of the subspace to achieve sufficient privacy preservation.
For example, it is common practice to $\ell_2$-normalize descriptors~\cite{Lowe2004Distinctive,Mishchuk2017Working,Liu2017SphereFace,Deng2018ArcFace}.
In such cases, lifting descriptors to affine lines ($m = 1$) is not secure. This is because a line intersects the unit hyper-sphere in at most $2$ points. It can be easy to detect which of the two intersections is statistically plausible and thereby exactly recover the original point.
However, any value of $m > 1$ generally produces infinite intersection points and thus provides much more ambiguity which is desirable for privacy preservation.
We now describe different lifting strategies, which we later compare in our experimental evaluation.

\customparagraph{Random Basis.}
One could sample random direction vectors for the linear subspace, \ie, $d_i \sim \mathcal{U}([-1, 1])^n$ referred to as \textit{random lifting}.
In our experiments, we found this approach to be vulnerable to relatively simple privacy attacks.
The original descriptor can be approximated by the nearest entry from a database of real-world descriptors according to the point-to-subspace distance.
This is because random subspaces generally intersect the descriptor manifold once.

\customparagraph{Adversarial Basis.} To address this issue, one can ensure that the subspace passes through multiple regions of the descriptor manifold.
We propose to use a database of real-world descriptors $W = \{w_1, \dots, w_s\}$ as an approximation of the manifold and sample the basis vectors as $d_i = w_i - d$, where $w_i \sim \mathcal{U}(W)$.
We call this approach \textit{adversarial lifting}, as it intentionally introduces plausible samples in the subspace to conceal the original descriptor.
Moreover, a defender can choose adversarial samples to hide specific private information, \eg, to hide the gender of a person, one can pick a feature vector from another gender, as shown in our experimental evaluation.
Adversarial sampling improves privacy but reduces descriptor matching accuracy, because the chance of accidental subspace intersections increases.
To balance the accuracy and privacy trade-offs, we propose combining the adversarial and random lifting strategies, which we call \textit{hybrid lifting}.
In hybrid lifting, a subset of the basis vectors are selected randomly while the rest are chosen using adversarial sampling.
There are different ways to implement the adversarial and hybrid strategies depending on the task at hand.
We describe a few such variants in the context of local features and face descriptors in Section~\ref{sec:experiments}.

\customparagraph{Translation Vector.} The origin can be set to any point in the subspace, except for the vector $d$ itself, since it is precisely what we must conceal.
Thus, we sample a random point and project it to the subspace, as follows:
\begin{equation}
	d_0 = p_\perp^{d + \vectorspan (d_1, \dots, d_m)} (e) \text{ where } e \sim \mathcal{U}([-1, 1])^n \enspace .
\end{equation}

\customparagraph{Information Leakage.} It is important to carefully construct the subspace to avoid accidental leakage of information.
For instance, in the adversarial formulation described above, all basis vectors ($d_i$ for $i>0$) point ``away from" the initial descriptor $d$.
An attacker could target parts of the descriptor manifold where these directions are feasible. 
More precisely, one could look for real-world descriptors $\tilde{d}$ such that $\tilde{d} + \lambda d_i$ also intersects the descriptor manifold.
To mitigate this, given an initial subspace $\mathcal{D}$, we generate a random basis as:
\begin{equation}
	d_i = p_\perp^{\mathcal{D}} (e_i) - d_0 \text{ where } e_i \sim \mathcal{U}([-1, 1])^n, \forall i \enspace .
\end{equation}
\section{Experimental Evaluation}
\label{sec:experiments}

In this section, we evaluate our method on two applications.
First, we experiment with local features on the task of image matching for visual localization and mapping.
Second, we apply our method to global image features for face verification.
We report results in these two settings and assess the trade-offs between the degree of privacy preservation achieved, the accuracy of the target task and the computational complexity.
As we cannot provide any theoretical guarantees on privacy preservation, we implement plausible privacy attacks and empirically demonstrate that our approach is robust against them.

\subsection{Runtime}

Previous approaches to privacy-preserving descriptors take advantage of homomorphic encryption.
While these methods guarantee an exact distance computation, they are severely limited in terms of practical applicability, especially in real-time scenarios.
A recent work about encrypted representation search~\cite{engelsma2020hers} reports that computing the distances between a single 128-dimensional query vector and a database with $1000$ entries takes around $1$ second (\cf~Figure~3~\cite{engelsma2020hers}).
Thus, obtaining the full distance matrix for an image pair with $1000$ features each would take around $16$ minutes.
In comparison, our method only induces minimal computational overhead, as shown in Table~\ref{tab:runtime}.
For completeness, the runtime for computing the point-to-point distance matrix in the same setting is $1.01\pm0.10$ms on GPU and $1.05\pm0.46$ms on CPU, respectively.

\setlength{\tabcolsep}{2.5pt}
\begin{table}
	\footnotesize
	\centering
	\begin{tabular}{c | c | c c c}
		\toprule
		\multirow{2}{*}{Dist.} & \multirow{2}{*}{Time (ms)} & \multicolumn{3}{c}{Subspace dimension} \\
		& & 2 & 4 & 8 \\ \midrule
		\multirow{2}{*}{s-to-s} & GPU & $2.02 \pm 0.14$ & $6.02 \pm 0.14$ & N/A \\
		& CPU & $107.87 \pm 0.95$ & $195.50 \pm 2.02$ & $540.98 \pm 25.18$ \\ \midrule
		\multirow{2}{*}{p-to-s} & GPU & $2.02 \pm 0.14$ & $2.10 \pm 0.30$ & $4.17 \pm 0.38$ \\
		& CPU & $25.25 \pm 1.14$ & $37.71 \pm 0.55$ & $63.24 \pm 1.08$ \\ \bottomrule
	\end{tabular}
	\caption{{\bf Runtime.} We report the average runtime over $100$ runs of the distance matrix computation for an image pair, when varying the lifting dimension. Each image has $1000$ $128$-dimensional floating point features. We consider both the subspace-to-subspace (s-to-s) and the point-to-subspace (p-to-s) distance. For the former, we implemented specialized CUDA solvers for lifting dimensions 2 and 4. Hardware: NVIDIA RTX 2080Ti, Intel Core i9-9900K.}
	\label{tab:runtime}
	\vspace{-5pt}
\end{table}
\setlength{\tabcolsep}{\tabcolsepdefault}

\begin{figure*}
	\centering
	\includegraphics[width=0.98\textwidth]{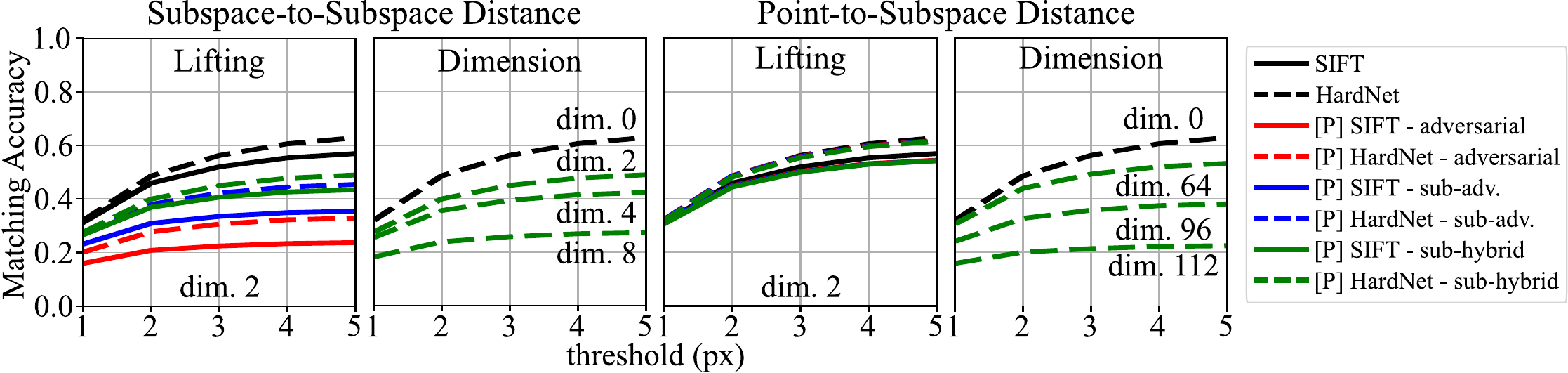}
	\vspace{-5pt}
	\caption{{\bf Matching evaluation.} We plot the mean matching accuracy at different thresholds on the HPatches sequences~\cite{Balntas2017HPatches}. Methods using our private representation are prefixed by {[P]}. We report results with different lifting methods and dimensions. HardNet outperforms SIFT on this benchmark and the ordering is respected after lifting as well.}
	\label{fig:hpatches}
	\vspace{-5pt}
\end{figure*}

\subsection{Local Feature Descriptors}

In order to demonstrate the robustness and generalizability of our approach, we perform experiments using the arguably most popular hand-crafted local feature (SIFT~\cite{Lowe2004Distinctive}) as well as a recent state-of-the-art learned descriptor (HardNet~\cite{Mishchuk2017Working}).
Both descriptors are by default $\ell_2$-normalized.
We evaluate the private descriptors on the tasks of image matching, structure-from-motion and visual localization.

\customparagraph{Subspace Selection.}
The adversarial lifting database is obtained by clustering 10 million local features from \numprint{60000} images of the Places365 dataset~\cite{Zhou2017Places} into $s=$ \numprint{256000} clusters using spherical k-means~\cite{Bishop2006Pattern}.	
In the context of 3D computer vision tasks, it is usually desirable to have many thousands of features per image~\cite{Schonberger2017Comparative}.	
Let us consider the case of lifting to descriptor planes ($m = 2$) using uniform random sampling from the database of \numprint{256000} centroids.	
Given an image pair with \numprint{8000} descriptors per image,	
for each feature in the second image, there is a $\nicefrac{1}{16}$ chance of sampling a feature already selected in the first one.	
Such a collision causes subspace intersections and thus leads to wrong feature matches.
This is further exacerbated by typical match filtering strategies (\eg,  mutual check, ratio-test~\cite{Lowe2004Distinctive}).

To reduce the number of wrong matches, we randomly split the database into $S$ pairwise disjoint sub-databases $W_1, \dots, W_S$ satisfying $W = \cup_{i=1}^S W_i, \card(W_i) = \nicefrac{s}{S}$.	
For an image $I$, we then first randomly select a sub-database $\mathcal{W} \sim \mathcal{U}(\{W_1, \dots, W_S\})$.	
Next, the basis vectors are generated using only elements of $\mathcal{W}$, \ie, $v_i = w_i - d$, where $w_i \sim \mathcal{U}(\mathcal{W})$.	
If two images select different sub-databases in this \textit{sub-adversarial lifting} strategy, the probability of random collision is $0$.	
For images using the same sub-database, the number of collisions is very high.	
Overall, with this strategy, instead of degrading the matching performance for all image pairs, we achieve good matching performance in $\nicefrac{15}{16}$ cases for $S=16$.	
In addition, we also evaluate a \textit{sub-hybrid lifting} strategy, where half of the basis vectors are random and the other half uses a sub-database.

\customparagraph{Image Matching.}
We compare raw descriptors with their private counterparts on the image sequences from the HPatches dataset~\cite{Balntas2017HPatches}.
This dataset consists of $116$ scenes with $6$ images each: $57$ of them exhibit illumination changes, while the other $59$ show significant viewpoint changes.
For each scene, we match the first image against the other $5$ yielding $580$ image pairs in total.
For evaluation, we follow protocol introduced by Dusmanu~\etal~\cite{Dusmanu2019D2} which reports the mean matching accuracy of a mutual nearest neighbors matcher for different values of the threshold up to which a match is considered correct.

Figure~\ref{fig:hpatches} shows results for both distances with different lifting methods and dimensions.
Random lifting is not plotted as it performs identical with the raw descriptors.
As mentioned above, adversarial lifting performs poorly for local features due to subspace collisions.
This is, in part, addressed by the use of sub-databases and further improved by sub-hybrid lifting.
The point-to-subspace distance only preserves the privacy of one image and is useful for cloud- and client-based visual localization systems, equivalent to Speciale~\etal~\cite{Speciale2019a,Speciale2019b}.
This approach is able to achieve good matching performance even for very high lifting dimensions.

\setlength{\tabcolsep}{2.5pt}
\begin{table*}
	\footnotesize
	\begin{minipage}{0.5\textwidth}
		\centering
		\begin{tabular}{c c c c c c}
			\toprule
			\multirow{2}{*}{Dataset} & \multirow{2}{*}{Method} & \multirowcell{2}{Reg.\\images} & \multirowcell{2}{Sparse\\points} & \multirowcell{2}{Track\\length} & \multirowcell{2}{Reproj.\\error} \\
			& & & & & \\ \midrule
			\multirowcell{6.5}{\textit{Madrid} \\ \textit{Metropolis} \\ $1344$ images} & SIFT & 400 & \numprint{28862} & 7.01 & 0.72 \\
			& [P] SIFT - dim. 2 & 302 & \numprint{17232} & 6.37 & 0.59 \\
			& [P] SIFT - dim. 4 & 227 & \numprint{11461} & 5.54 & 0.56 \\ \cmidrule{2-6}
			& HardNet & 459 & \numprint{42180} & 7.25 & 0.89 \\
			& [P] HardNet - dim. 2 & 367 & \numprint{28367} & 6.49 & 0.68 \\
			& [P] HardNet - dim. 4 & 268 & \numprint{15562} & 6.32 & 0.58 \\ \midrule
			\multirowcell{6.5}{\textit{Gendarmen-} \\ \textit{markt} \\ $1463$ images} & SIFT & 896 & \numprint{74348} & 6.37 & 0.84 \\
			& [P] SIFT - dim. 2 & 783 & \numprint{64554} & 5.44 & 0.71 \\
			& [P] SIFT - dim. 4 & 458 & \numprint{33291} & 5.23 &  0.60 \\ \cmidrule{2-6}
			& HardNet & 999 & \numprint{112245} & 6.68 & 0.96 \\
			& [P] HardNet - dim. 2 & 864 & \numprint{89865} & 5.98 & 0.80 \\
			& [P] HardNet - dim. 4 & 751 & \numprint{63862} & 5.50 & 0.69 \\ \midrule
			\multirowcell{6.5}{\textit{Tower of} \\ \textit{London} \\ $1576$ images} & SIFT & 635 & \numprint{64490} & 7.78 & 0.70 \\
			& [P] SIFT - dim. 2 & 525 & \numprint{55439} & 6.58 & 0.61 \\
			& [P] SIFT - dim. 4 & 439 & \numprint{37819} & 6.10 & 0.56 \\ \cmidrule{2-6}
			& HardNet & 749 & \numprint{89818} & 7.85 & 0.81 \\
			& [P] HardNet - dim. 2 & 557 & \numprint{69161} & 7.19 & 0.68 \\
			& [P] HardNet - dim. 4 & 498 & \numprint{49570} & 6.69 & 0.61 \\
			\bottomrule
		\end{tabular}
	\end{minipage}
	\hfill
	\begin{minipage}{0.16\textwidth}
		\centering
		{\normalsize Madrid Metropolis}\\
		HardNet\\
		\includegraphics[height=.75\textwidth]{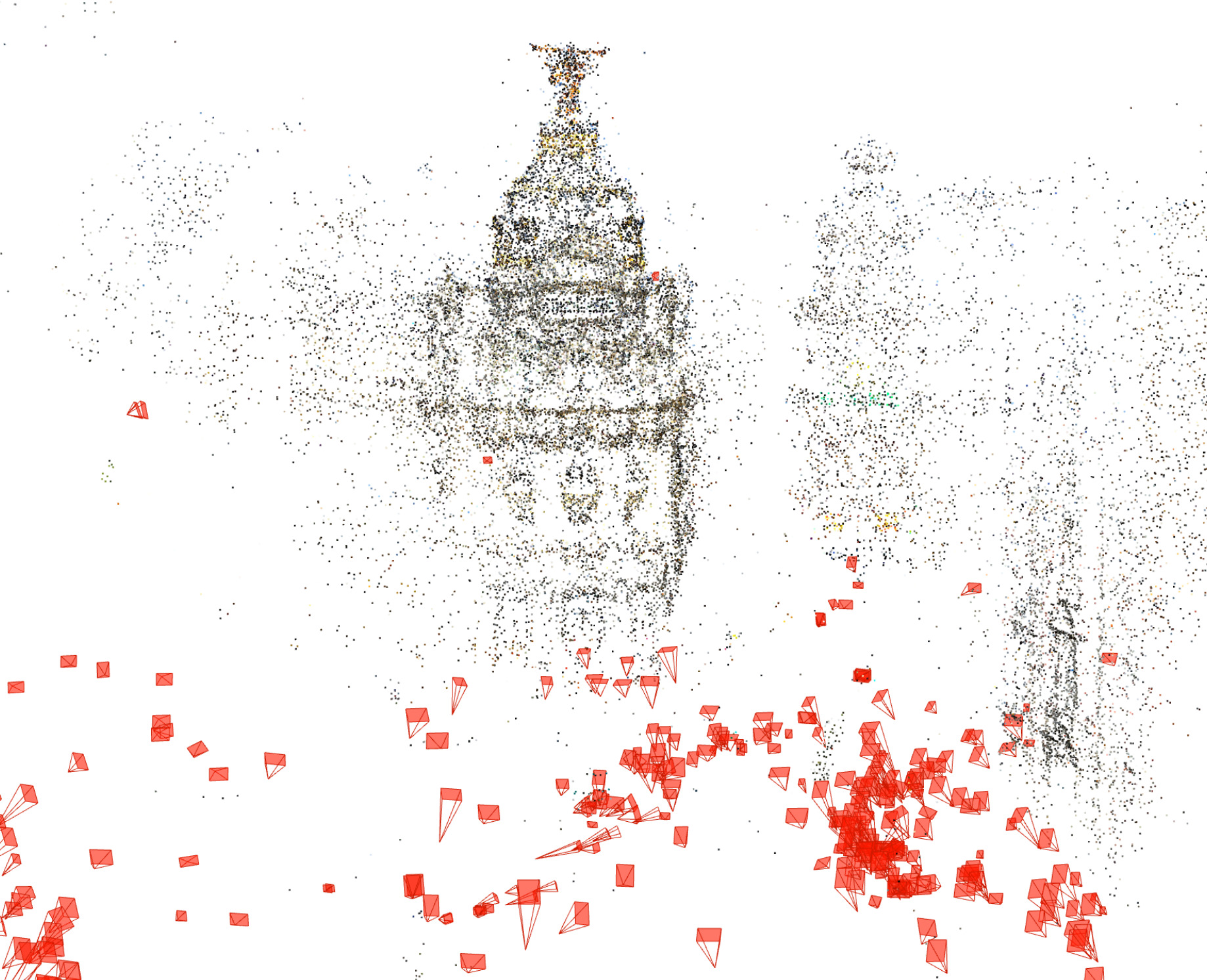}\\
		{[P]} HardNet - dim. 2\\
		\includegraphics[height=.75\textwidth]{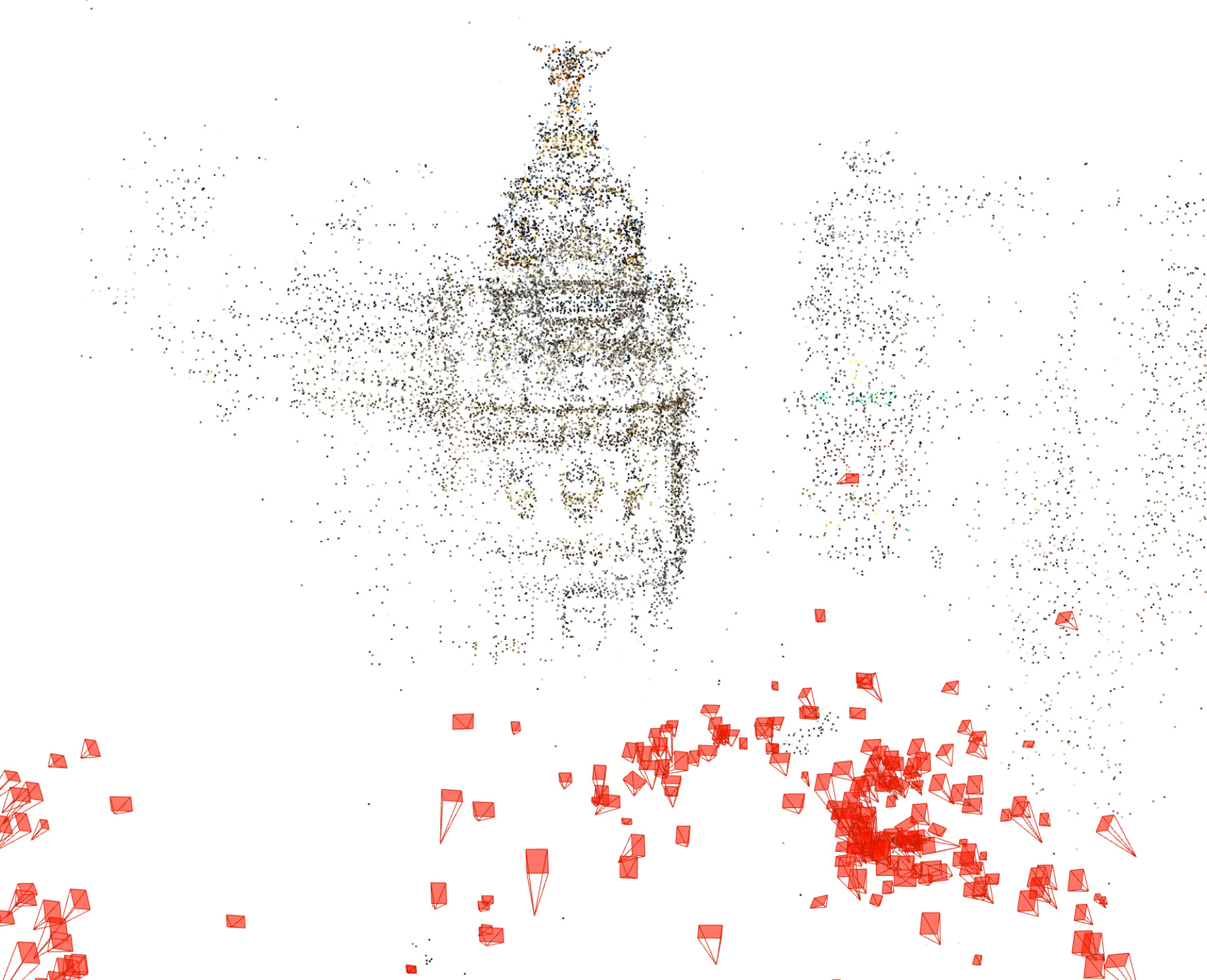}\\
		{[P]} HardNet - dim. 4\\
		\includegraphics[height=.75\textwidth]{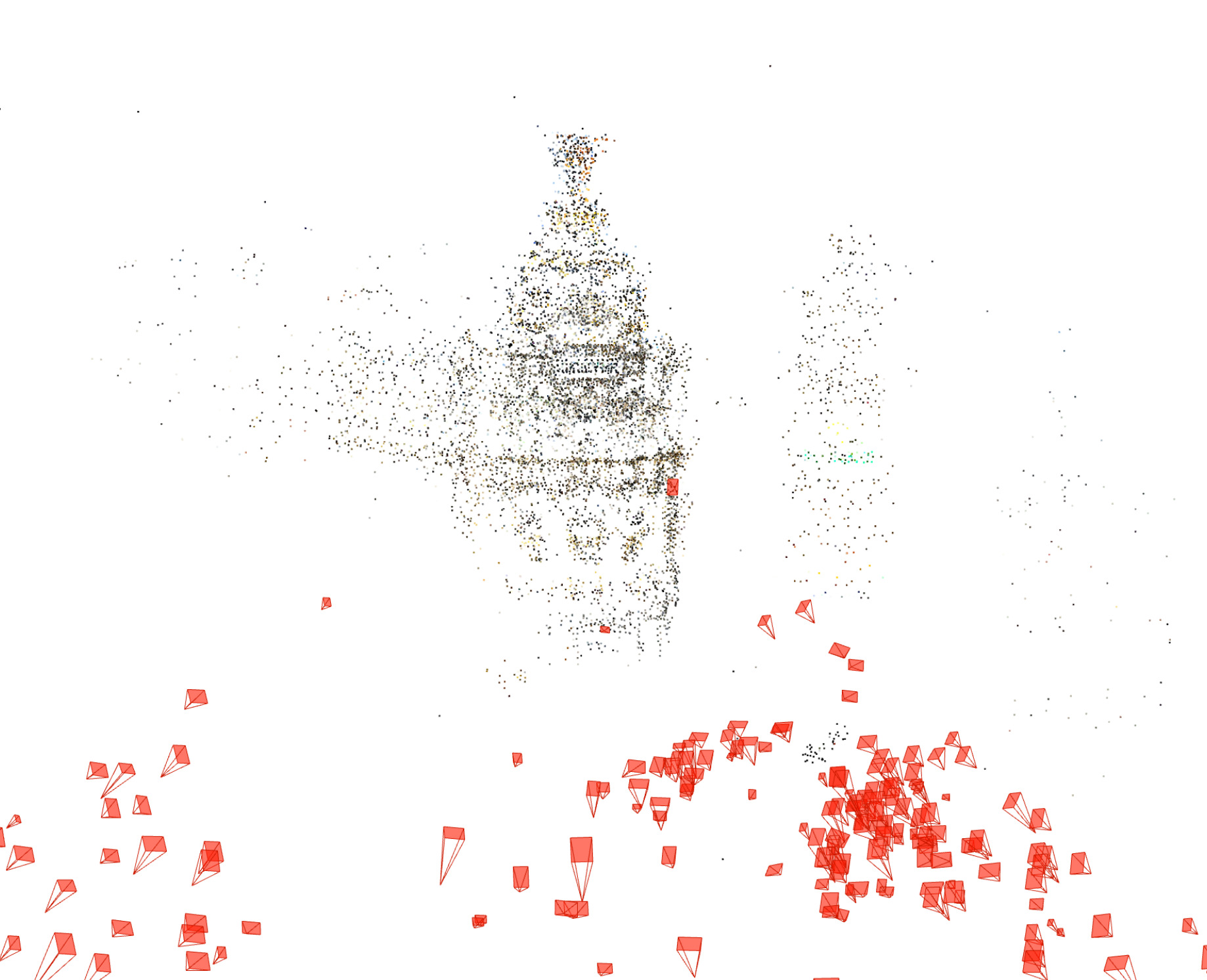}
	\end{minipage}
	\hfill
	\begin{minipage}{0.32\textwidth}
		\centering
		{\normalsize Tower of London}\\
		HardNet\\
		\includegraphics[height=.375\textwidth]{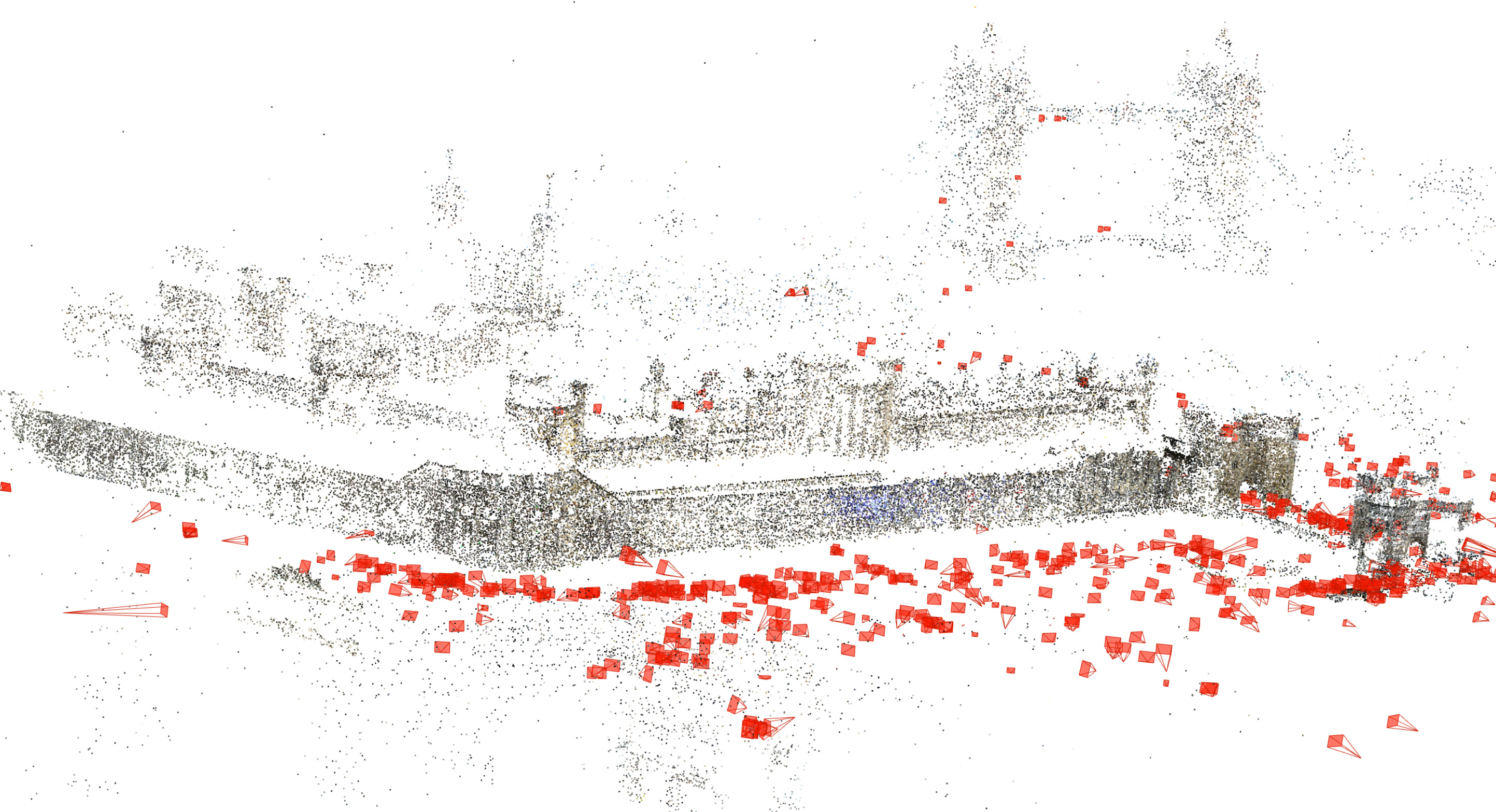}\\
		{[P]} HardNet - dim. 2\\
		\includegraphics[height=.375\textwidth]{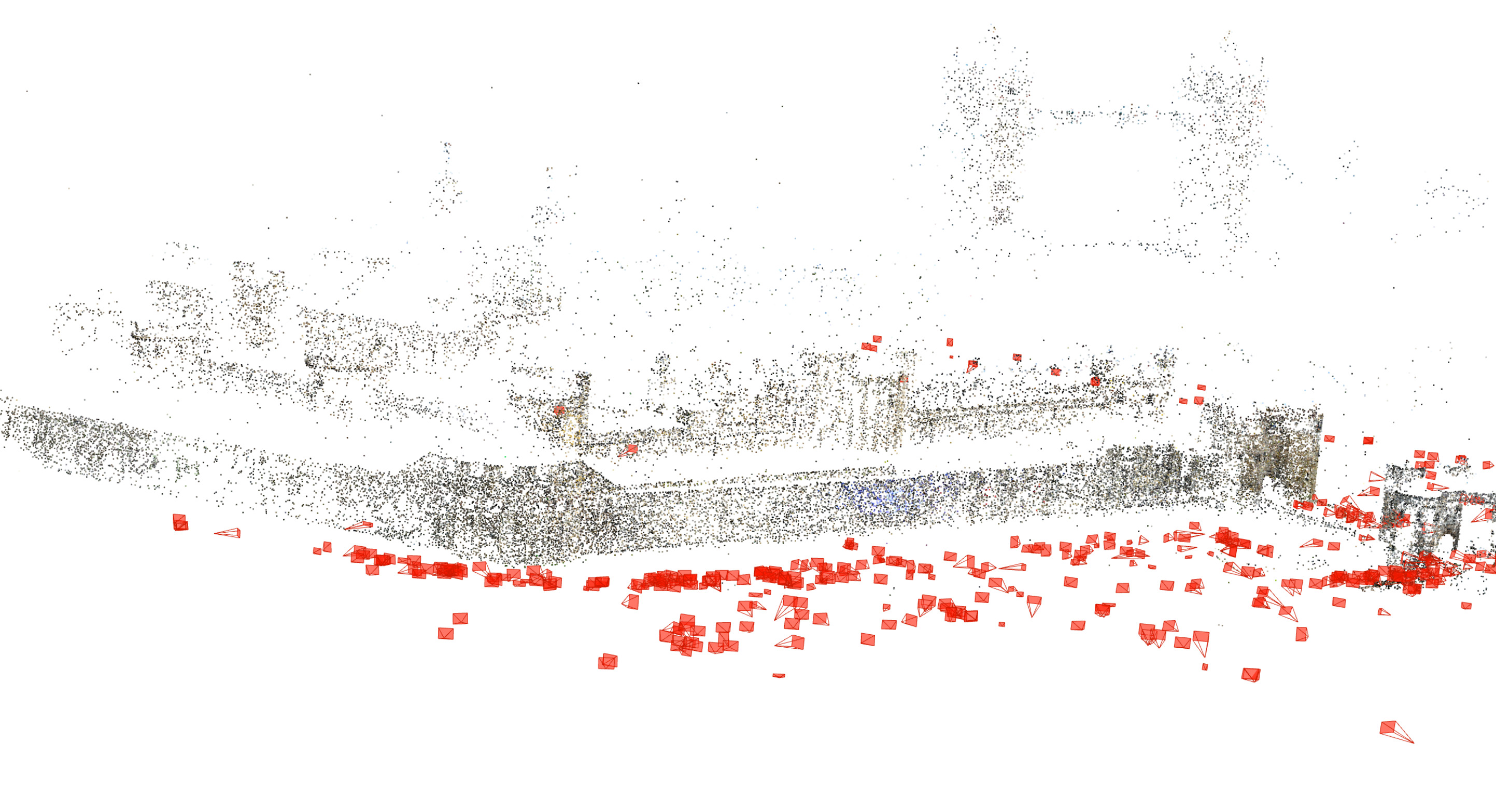}\\
		{[P]} HardNet - dim. 4\\
		\includegraphics[height=.375\textwidth]{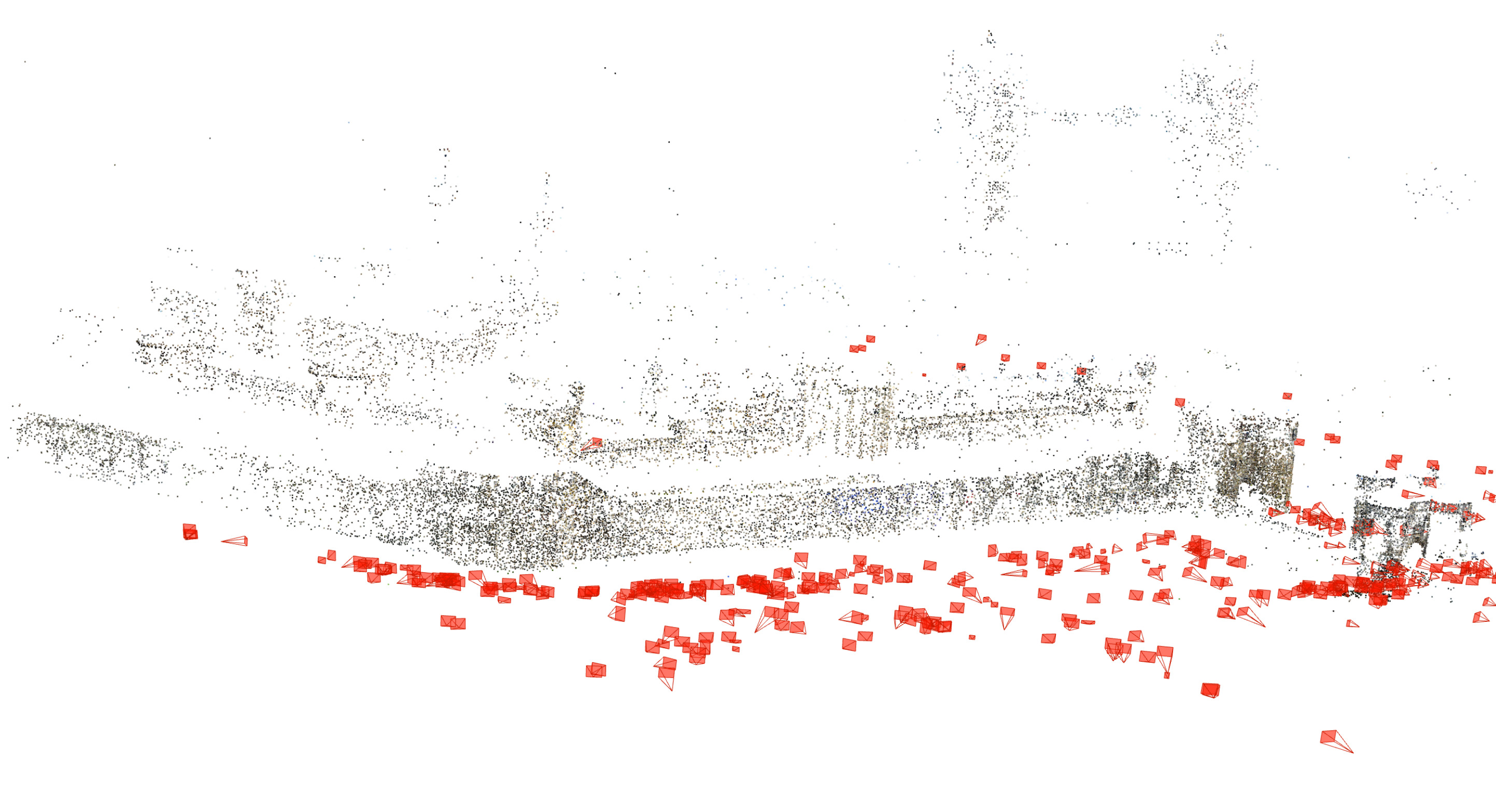}
	\end{minipage}
	\caption{{\bf Local Feature Evaluation Benchmark.} We report reconstruction statistics such as the number of registered images and sparse points and the average track length and reprojection error on internet photo collections of landmarks~\cite{Schonberger2017Comparative}. Methods prefixed by {[P]} use sub-hybrid lifting for all features of input images. On the right side, we visualize the final sparse models.}
	\label{tab:lfe}
	\vspace{-5pt}
\end{table*}
\setlength{\tabcolsep}{\tabcolsepdefault}

\customparagraph{Structure-from-Motion.}
Next, we integrate the best performing private representation from above (sub-hybrid lifting) into an end-to-end 3D reconstruction pipeline~\cite{Schoenberger2016Structure} and evaluate it on the crowd-sourced 3D reconstruction benchmark of Sch\"onberger~\etal~\cite{Schonberger2017Comparative}.
For each image, we retrieve the top $50$ most similar images using NetVLAD~\cite{Arandjelovic2016NetVLAD} and only match against these.
Next, we run geometric verification (with a minimum inlier ratio of $0.1$) followed by sparse reconstruction using COLMAP~\cite{Schoenberger2016Structure,schoenberger2016vote} and finally report the reconstruction statistics in Table~\ref{tab:lfe}. For this evaluation, we preserve the privacy of all input images.
As already observed in our image matching evaluation, the private features come with accuracy trade-offs.
As we increase the dimensionality of the subspace, the reconstruction completeness degrades accordingly.
Despite the fewer number of registered images, the 3D models remain relatively accurate and clearly distinguishable.
The generally lower track length for private features is caused by missing matches leading to longer feature tracks being split into multiple smaller ones.

\setlength{\tabcolsep}{2.5pt}
\begin{table}
	\footnotesize
	\centering
	\begin{tabular}{c c c c c}
		\toprule
		\multirow{2}{*}{Query} & \multirow{2}{*}{Method} & \multicolumn{3}{c}{Thresholds} \\
		& & $0.25$m, $2^{\circ}$ & $0.5$m, $5^{\circ}$ & $5.0$m, $10^{\circ}$ \\ \midrule
		\multirowcell{8.5}{Day\\($824$)} & SIFT & $82.9\%$ & $89.6\%$ & $92.2\%$ \\
		& [P] SIFT - dim. 2 & $79.5\%$ & $87.0\%$ & $91.1\%$ \\
		& [P] SIFT - dim. 4 & $79.6\%$ & $86.5\%$ & $91.1\%$ \\
		& [P] SIFT - dim. 16 & $76.7\%$ & $84.0\%$ & $87.4\%$ \\
		\cmidrule{2-5}
		& HardNet & $86.3\%$ & $92.5\%$ & $95.6\%$ \\
		& [P] HardNet - dim. 2 & $84.3\%$ & $89.8\%$ & $94.3\%$ \\
		& [P] HardNet - dim. 4 & $83.5\%$ & $90.2\%$ & $93.6\%$ \\
		& [P] HardNet - dim. 16 & $82.0\%$ & $88.3\%$ & $92.2\%$ \\
		\midrule
		\multirowcell{8.5}{Night\\($98$)} & SIFT & $41.8\%$ & $48.0\%$ & $55.1\%$ \\
		& [P] SIFT - dim. 2 & $32.7\%$ & $36.7\%$ & $42.9\%$ \\
		& [P] SIFT - dim. 4 & $32.7\%$ & $38.8\%$ & $43.9\%$ \\ 
		& [P] SIFT - dim. 16 & $25.5\%$ & $31.6\%$ & $34.7\%$ \\
		\cmidrule{2-5}
		& HardNet & $60.2\%$ & $67.3\%$ & $73.5\%$ \\
		& [P] HardNet - dim. 2 & $49.0\%$ & $53.1\%$ & $58.2\%$ \\
		& [P] HardNet - dim. 4 & $40.8\%$ & $44.9\%$ & $49.0\%$ \\
		& [P] HardNet - dim. 16 & $32.7\%$ & $37.8\%$ & $43.9\%$ \\
		\bottomrule
	\end{tabular}
	\caption{{\bf Aachen Day-Night Localization Challenge.} We report the percentage of localized query images for both day and night scenarios under different camera pose accuracy threshold on the Aachen Day-Night dataset~\cite{Sattler2017Benchmarking}. For the private methods (prefixed by {[P]}), we use sub-hybrid lifting for query images and point-to-subspace distance for matching.}
	\label{tab:aachen}
	\vspace{-5pt}
\end{table}
\setlength{\tabcolsep}{\tabcolsepdefault}

\customparagraph{Visual Localization.}
We also consider the case of localizing to an already built map on the challenging Aachen Day-Night long-term visual localization dataset~\cite{Sattler2017Benchmarking}.
This is equivalent to the scenario tackled by Speciale~\etal~\cite{Speciale2019b}, where the goal is to protect the privacy of users of an image-based localization service, such as Google Visual Positioning System~\cite{Google2019Google} or Microsoft Azure Spatial Anchors~\cite{Microsoft2019Announcing}.
We first triangulate the database model from the given camera poses and intrinsics using DoG keypoints with raw SIFT and HardNet descriptors, respectively.
For each query image ($824$ day-time and $98$ night-time), we retrieve the top $50$ database images using NetVLAD~\cite{Arandjelovic2016NetVLAD}.
We preserve the privacy of all query images with sub-hybrid lifting and use point-to-subspace distance for matching.
Finally, we use the COLMAP~\cite{Schoenberger2016Structure} image registrator with fixed intrinsics to obtain poses that are submitted to the long-term visual localization benchmark~\cite{VisualLocalization}.

Following the standard evaluation protocol, we report the percentage of localized query images for different real-world thresholds in Table~\ref{tab:aachen}.
On the day queries, we are able to achieve competitive performance even when lifting 
to 16 dimensional subspaces.
As previously, the accuracy gradually decreases when increasing the lifting dimension.
Furthermore, even on the extremely hard night-to-day matching queries where pose estimation has very low inlier ratios, we are still able to localize a reasonable number of queries.

\setlength{\tabcolsep}{2.5pt}
\begin{figure*}
	\centering
	\begin{subfigure}[c]{.30\textwidth}
		\footnotesize
		\centering
		\begin{tabular}{c c c c c c}
			\toprule
			\multirowcell{3}{\rotatebox[origin=c]{90}{Attack}} & \multirowcell{3}{Lifting} & \multirowcell{3}{Dim.} & \multirowcell{3}{MAE\\$(\downarrow)$} & \multirowcell{3}{SSIM\\$(\uparrow)$} & \multirowcell{3}{PSNR\\$(\uparrow)$} \\ 
			& & & & & \\
			& & & & & \\ \midrule
			& raw & 0 & 0.105 & 0.755 & 17.937 \\ \midrule
			\multirowcell{2}{\rotatebox[origin=c]{90}{NNA}} & random & 2 & 0.112 & 0.738 & 17.448 \\ 
			& sub-hybrid & 2 & 0.206 & 0.530 & 12.288 \\ \midrule
			\multirowcell{3}{\rotatebox[origin=c]{90}{DIA}} & \multirowcell{3}{sub-hybrid} & 2 & 0.176 & 0.648 & 13.447 \\
			& & 4 & 0.179 & 0.594 & 13.531 \\
			& & 6 & 0.194 & 0.559 & 12.823 \\
			\bottomrule
		\end{tabular}
	\end{subfigure}
	\hfill
	\begin{subfigure}[c]{.65\textwidth}
		\centering
		\includegraphics[width=\textwidth]{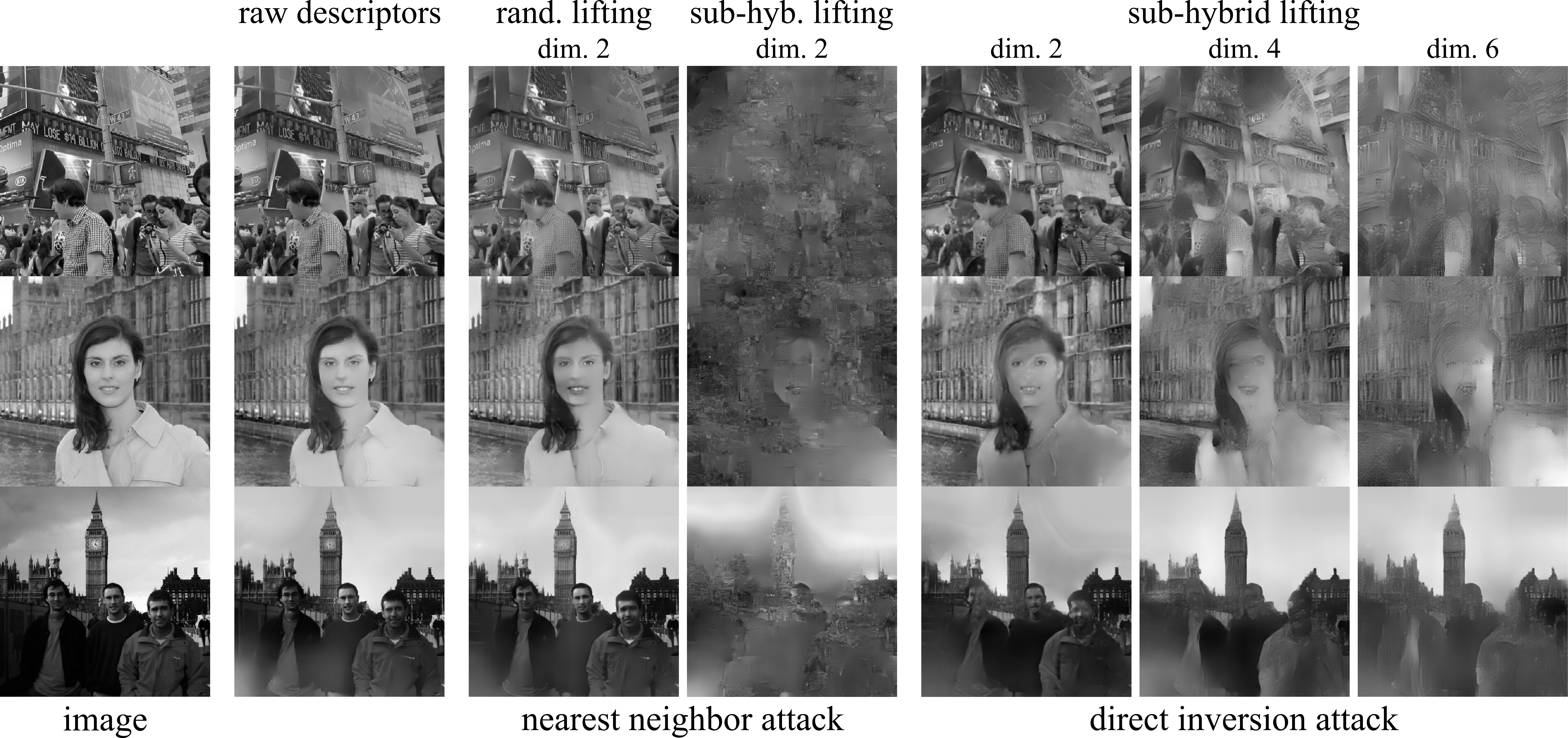}
	\end{subfigure}
	\vspace{-10pt}
	\caption{{\bf Image reconstruction.} On the left, we report quality metrics between reconstructed and original images. On the right, we show several qualitative examples: first original image, then reconstructions from the raw descriptors and using the proposed privacy attacks on different lifting methods and dimensions.  Image credit (top to bottom): \textit{pagedooley} (Kevin Dooley), \textit{laylamoran4battersea} (Layla Moran), \textit{martinalvarez} (Martin Alvarez Espinar).}
	\label{fig:image_rec}
	\vspace{-5pt}
\end{figure*}
\setlength{\tabcolsep}{\tabcolsepdefault}

\customparagraph{Privacy Attack.}
To analyze attacks on the proposed private descriptor representation, we provide the adversary with multiple tools.
We assume that they have access to a database $V$ of \numprint{128000} real-world descriptors built using the same procedure as the lifting database (described above).
Further, the attacker has unrestricted access to the lifting algorithm and is able to use it on-demand.
Finally, they have access to extensive training data (the MegaDepth~\cite{Li2018MegaDepth} dataset) as well as the architecture and loss from Pittaluga~\etal~\cite{Pittaluga2019} allowing them to train new feature inversion networks.

First, we consider a nearest neighbor attack (NNA) where each subspace is approximated by its closest correspondence from a database of real-world descriptors.
Formally, for each private representation $\mathcal{D}$ associated to a descriptor $d$, the database $V$ is used to retrieve the closest element to the subspace $\tilde{d} = \arg\min_{v \in V} \text{dist}(\mathcal{D}, v)$.
Next, the approximated descriptors $\tilde{d}$ can be fed to a regular feature inversion network to reconstruct the appearance of the original image.

Second, we consider a direct inversion attack (DIA) where the affine subspaces are provided as input to a CNN.
To this end, we train multiple feature inversion networks from Difference-of-Gaussians (DoG) keypoints and private descriptors lifted to 2, 4, and 6 dimensions, respectively.
Note that the architectures proposed in previous works~\cite{Pittaluga2019,Dosovitskiy2016Inverting} are very compute and memory intensive -- training them on higher dimensional subspaces would be a challenge in itself.

We run the proposed privacy attacks on $10$ images\footnote{We manually selected $2$ holiday images from Hong Kong, London, New York, Paris, and Tokyo published on Flickr under a \href{https://creativecommons.org/licenses/by/4.0/}{CC BY 4.0 License}.} using HardNet descriptors and present the results in Figure~\ref{fig:image_rec}.
On the left, we quantitatively report image reconstruction quality metrics such as mean absolute error (MAE), structural similarity index measure (SSIM), and peak signal-to-noise ratio (PSNR); on the right, we show qualitative image reconstructions.
Please refer to the supplementary material for more examples.
Using the raw descriptors, one can reconstruct the original image with very high fidelity (note the readability of text in the first example).
The nearest neighbor attack is successful on private features using random lifting, but not when using sub-hybrid lifting due to the adversarial samples.
For all reconstructions, the general outline of the buildings is recovered mainly due to the lack of features in the sky (\eg, third example).
The direct inversion attack is able to reconstruct some parts of the original image, but the quality is significantly deteriorated.
Furthermore, distinguishing details such as faces or text are heavily perturbed and become non-existent for higher lifting dimensions.

\subsection{Face Descriptors}

For this evaluation, we use a state-of-the-art deep face descriptor -- the best performing ArcFace~\cite{Deng2018ArcFace} model with a ResNet-101~\cite{He2016Deep} backbone trained on MS-Celeb-1M~\cite{Guo2016MS}.

\begin{figure*}
	\centering
	\includegraphics[width=\textwidth]{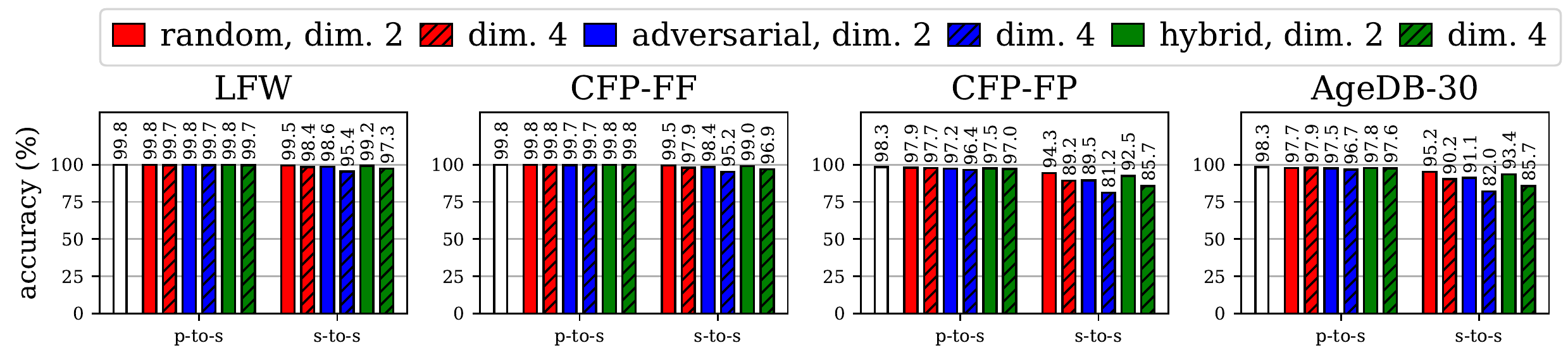}
	\vspace{-25pt}
	\caption{{\bf Face verification.} We show the accuracy on different face verification datasets. The white bar represents the reference accuracy of raw ArcFace descriptors. The point-to-subspace distance (\textit{p-to-s}), performs within at most $2\%$ of the original descriptors. For the subspace-to-subspace distance (\textit{s-to-s}), the performance drop is more significant in the difficult scenarios (CFP-FP and AgeDB-30), but frontal authentication (LFW and CFP-FF) is still very accurate ($95\%$ at worst).}
	\label{fig:face_verif}
	\vspace{-5pt}
\end{figure*}

\customparagraph{Face Verification.}
We report face verification accuracy on multiple datasets: LFW~\cite{Huang2007Labeled}, CFP~\cite{Sengupta2016Frontal} (both frontal-frontal denoted FF and frontal-profile denoted FP), and AgeDB-30~\cite{Moschoglou2017AgeDB}.
We follow the regular evaluation protocol, notably $10$-fold cross validation where, for each fold, the training split is used to determine a distance threshold that separates between same / different identity and the accuracy is computed on the validation split. Finally, the mean classification accuracy over the $10$ folds is reported in Figure~\ref{fig:face_verif}.

We evaluate two scenarios: point-to-subspace (\textit{p-to-s}) matching, where one of the images is represented using the original descriptor and the other one is lifted to a subspace, and subspace-to-subspace (\textit{s-to-s}) matching, where both descriptors are private.
As expected, the point-to-subspace matching performs better across the board.
For the subspace-to-subspace distance, the performance on the simple datasets (LFW and CFP-FF) only drops by a few percents.
For more complex datasets (frontal-profile matching in CFP-FP, large age differences in AgeDB-30), the performance drop is more significant.
Nevertheless, the simpler datasets are still very representative of common authentication systems (Microsoft Windows Hello~\cite{WindowsHello}, Apple Face ID~\cite{FaceID}), making our approach highly relevant for such scenarios.

\begin{figure*}
	\centering
	\includegraphics[width=\textwidth]{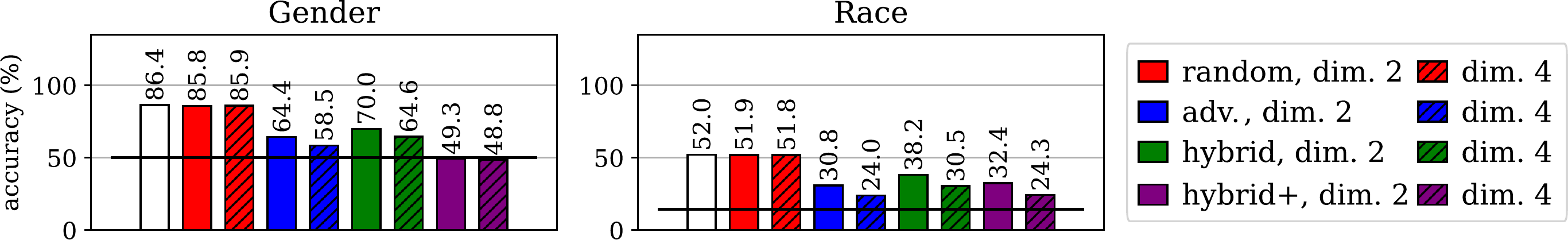}
	\vspace{-20pt}
	\caption{{\bf FairFace.}  We report the accuracy of a K-NN classifier trained to predict the gender and race of a subject from their ArcFace descriptor. The black line represents the approximate accuracy of a random classifier. The white bars represent the accuracy on raw ArcFace descriptors. Private representations using a database for lifting successfully conceal information.}
	\label{fig:fairface}
	\vspace{-5pt}
\end{figure*} 

\customparagraph{Privacy Attack.}
The privacy attack we are concerned with involves inferring distinguishing properties (gender, race) from only the ArcFace~\cite{Deng2018ArcFace} descriptors.
For this purpose, we used FairFace~\cite{Karkkainen2019Fair}, a face dataset consisting of \numprint{97698} images with balanced gender (2 classes) and race (7 classes) annotations.
We randomly selected \numprint{10000} training images for the database needed by our lifting method.
The remaining \numprint{76744} training images were used for the attack.
The validation set of \numprint{10954} images is used for evaluation.

We attack an ArcFace descriptor using a K-nearest neighbors (K-NN) classifier~\cite{Bishop2006Pattern} to predict the gender and race of the person.
We do this both on the original feature as well as the lifted feature for $K=10$.
We also implemented a variant of our hybrid lifting method (denoted \textit{hybrid+}) that exploits the gender / race of each person.
In this variant, each feature is lifted by sampling database entries with a different gender / of a different race to obtain a balanced subspace, which better conceals these attributes.

The results are reported in Figure~\ref{fig:fairface}.
The black vertical lines denote the approximate performance of a random classifier.
Similar to image matching, pure random lifting is again not effective at concealing the private attributes.
Adversarial lifting has the best results in terms of privacy, but its face verification accuracy is also the worst.
Hybrid lifting offers a trade-off between random lifting (high performance) and adversarial lifting (good for privacy).
Finally, the hybrid+ version is most effective at concealing the gender.
\section{Limitations and Future Work}
Speciale \etal~\cite{Speciale2019b} showed that solving the target task of camera localization reveals the concealed location of some features in the query image.
Similarly, in our 3D reconstruction task, the pair of closest points on two matched affine subspaces provides a way to estimate the concealed feature descriptors.
This implies that features associated with 3D points triangulated from multiple views are likely to be revealed.
By inverting the estimated descriptors, an adversary might be able to approximately reconstruct the appearance of the stationary part of the scene.
However, this is not a serious limitation, as feature descriptors extracted from image regions depicting people or other transient objects will generally not be matched in multiple overlapping images and therefore their appearance is unlikely to be revealed.

For face verification, it is possible to infer the face descriptor after repeated authentications of a person if a history of the private descriptors is stored.
One potential mitigation is to generate near parallel subspaces for a particular individual, although it is unclear how this approach behaves with respect to the manifold of face descriptors.
A potential option would be adding a trusted third-party in the system that receives private descriptors from both client and server and computes the distances without storing any data.

Apart from addressing these limitations, other directions for future work include training descriptors more suitable for lifting and implementing scalable matching inspired by prior work on subspace representations~\cite{Basri2007Approximate} to enable large-scale applications such as place recognition.

\section{Conclusion}
\label{sec:conclusions}

We have proposed a novel privacy-preserving feature representation by embedding feature descriptors into affine subspaces containing adversarial samples.
To find similar features, nearest neighbor computation is enabled through point-to-subspace or subspace-to-subspace distance.
We experimentally demonstrate the high practical relevance of our approach for crowd-sourced visual localization and mapping as well as face authentication, while rendering it difficult to recover sensitive information.

{\noindent \textbf{Acknowledgements.} This work was supported by the Microsoft Mixed Reality \& AI Z\"urich Lab PhD scholarship.}

\appendix
\part*{Supplementary Material}
This document contains the following supplementary information.
First, we describe the dual formulation for the point-to-subspace and subspace-to-subspace distances.
Next, we discuss the space and time complexity of our matching algorithm.
Finally, we show more quantitative and qualitative results of privacy attacks on local features in the scenario of an image-based localization service.
\section{Dual Formulation}
\label{sec:dual}

Alternative to our formulation in the main paper, an $m$-dimensional linear subspace of $\mathbb{R}^n$ can also be interpreted as the intersection of $n - m$ hyperplanes. Under this formulation, an affine subspace can be defined by the sum of a translation vector $a_0$ and the orthogonal subspace of the linear span of $a_1, \dots, a_{n - m}$, \ie, $\mathcal{A} = a_0 + \vectorspan (a_1, \dots, a_{n - m})^\perp$. Throughout the entire section, we suppose that $(a_1, \dots, a_{n - m})$ is orthonormal, \ie, that $A = [a_1 \dots a_{n - m}]^T$ satisfies $A A^T = I$.

We consider two affine subspaces $\mathcal{D}, \mathcal{E}$ under this representation. Let $(x^*, y^*) \in \mathcal{D} \times \mathcal{E}$ be a solution of the subspace-to-subspace distance, \ie, $\lVert y^* - x^* \rVert = \min_{x \in \mathcal{D}, y \in \mathcal{E}} \lVert y - x \rVert$.
As before, a sufficient and necessary condition for $\text{dist}(\mathcal{D}, \mathcal{E}) = \lVert y^* - x^* \rVert$ is that the line $y^* - x^*$ is orthogonal to both $\mathcal{D}$ and $\mathcal{E}$, \ie, there exist $\boldsymbol{\mu}, \boldsymbol{\nu} \in \mathbb{R}^{n - m}$ such that $y^* - x^* = \sum_{j = 1}^{n - m} \mu_j d_j = \sum_{j = 1}^{n - m} \nu_j e_j$.
Finally, $x^*, y^*, \boldsymbol{\mu}, \boldsymbol{\nu}$ must satisfy the following constraints:
\begin{equation}
\begin{cases}
y^* - x^* = \sum_{j = 1}^{n - m} \mu_j d_j = \sum_{j = 1}^{n - m} \nu_j e_j \\
d_i^T (x^* - d_0) = 0 \\
e_i^T (y^* - e_0) = 0
\end{cases}
\end{equation}
which can be rewritten as:
\begin{equation}
\begin{cases}
\sum_{j = 1}^{n - m} \mu_j d_j - \sum_{j = 1}^{n - m} \nu_j e_j = \mathbf{0} \\
d_i^T x^* = d_i^T d_0 \\
e_i^T x^* + \sum_{j = 1}^{n - m} \nu_j e_i^T e_j = e_i^T e_0
\end{cases} \enspace .
\end{equation}
This system can be represented under the following form:
\begin{equation}
\begin{bmatrix}
D & \mathbf{0}_{(n - m)^2} & \mathbf{0}_{(n - m)^2} \\
E & \mathbf{0}_{(n - m)^2} & I \\
\mathbf{0}_{n^2} & D^T & -E^T
\end{bmatrix}
\begin{bmatrix}
x^* \\
\boldsymbol{\mu} \\
\boldsymbol{\nu}
\end{bmatrix} =
\begin{bmatrix}
D d_0 \\
E e_0 \\
\mathbf{0}
\end{bmatrix} \enspace ,
\end{equation}
where $D = \begin{bmatrix} d_1 \dots d_{n - m}\end{bmatrix}^T, E = \begin{bmatrix} e_1 \dots e_{n - m}\end{bmatrix}^T \in M_{(n - m) \times n}(\mathbb{R})$.

In this case, finding the subspace-to-subspace distance can be reduced to solving a linear system with $3n - 2m$ unknowns and equations.
This formulation is thus preferable when $m > \frac{3}{4} n$.

For the point-to-subspace distance between a private descriptor under this representation $\mathcal{D}$ and an original descriptor $e$, the system can be simplified to:
\begin{align}
& \begin{cases}
	e - x^* = \sum_{j=1}^{n - m} \mu_j d_j \\
	d_i^T (x^* - d_0) = 0
\end{cases} \\
\Leftrightarrow & \sum_{j = 1}^{n - m} \mu_j d_i^T d_j = d_i^T (e - d_0) \\
\Leftrightarrow & \mu_i = d_i^T (e - d_0) \enspace,
\end{align}
since $D D^T = I$.
Thus,
\begin{align}
	\text{dist}(\mathcal{D}, e) &= \lVert \sum_{j = 1}^{n - m} d_j^T (e - d_0) d_j \rVert \\
	& = \lVert p_\perp^{\vectorspan (d_1, \dots, d_{n - m})}(e - d_0) \rVert \enspace .
\end{align}
This formulation is more advantageous when $m \geq \frac{1}{2}n$ as it only requires $n - m$ dot product evaluations instead of $m$.
\section{Complexity Analysis}
\label{sec:complexity}

\customparagraph{Time Complexity.}
The complexity of lifting to an m-dimensional subspace is $\mathcal{O}(m n)$ under the supposition that the lifting database offers $\mathcal{O}(1)$ access to a random element (\eg, array, hashtable). 

In general, for matching two features lifted to $m$-dimensional affine subspaces under the primal representation, we require a matrix multiplication $(m \times n) (n \times m)$ (\ie, $M = -D E^T$), the resolution of a system with $2m$ unknowns and equations, and a constant number of additional matrix multiplications between $m \times m$ matrices.
Thus, the complexity is $\mathcal{O}(m^2 n + m^3)$.
Similarly, for the dual representation, the complexity is $\mathcal{O}((3n - 2m)^3)$.

To match two images with $N_1$ and $N_2$ local features respectively, we use exhaustive matching which requires computing distances between all pairs of features, \ie, a time complexity of $\mathcal{O}(N_1 N_2 C)$, where $C$ is the complexity of matching two features as defined above.

\customparagraph{Space Complexity.}
For the primal representation, we require one translation vector and $m$ basis vectors totaling $\mathcal{O}((m + 1) n)$ floating point variables instead of $\mathcal{O}(n)$ for the original features.
For the dual representation, we require storing $\mathcal{O}((n - m + 1) n)$ floating point variables.

\section{Privacy Attacks on Local Features}
\label{sec:attack}

In this section, we first provide additional results of the proposed privacy attacks on local features.
We then study a new oracle based attack underlining the effectiveness of adversarial lifting.

\customparagraph{Additional Results.}
We run the privacy attacks described in Section~4.2, paragraph {\it Privacy Attack} of the main paper on both SIFT and HardNet private features with different lifting methods and dimensions.
To recall, we proposed a nearest neighbor (NNA) and a direct inversion attack (DIA).
In Table~\ref{tab:image_rec_sift_stat}, we quantitatively report image reconstruction quality metrics such as mean absolute error (MAE), structural similarity
index measure (SSIM), and peak signal-to-noise ratio (PSNR) for SIFT descriptors.
We show additional qualitative results of the attack for SIFT and HardNet descriptors in Figures~\ref{fig:image_rec_qualitative_sift} and~\ref{fig:image_rec_qualitative_hardnet}, respectively.
All images were published on Flickr under a \href{https://creativecommons.org/licenses/by/4.0/}{CC BY 4.0 License}.
Image credit (top-to-bottom): \texttt{twang\_dunga} (Twang Dunga), \texttt{scaredykat} (krista), \texttt{bab4lity} (wwikgren), \texttt{herry} (Herry Lawford), \texttt{smemon} (Sean MacEntee), \texttt{laylamoran4battersea} (Layla Moran), \texttt{shankaronline} (Shankar S.), \texttt{martinalvarez} (Martin Alvarez Espinar), \texttt{pagedooley} (Kevin Dooley), \texttt{nukeit1} (James McCauley).

\setlength{\tabcolsep}{2.5pt}
\begin{table}
	\footnotesize
	\centering
	\begin{tabular}{c c c c c c}
		\toprule
		\multirowcell{3}{\rotatebox[origin=c]{90}{Attack}} & \multirowcell{3}{Lifting} & \multirowcell{3}{Dim.} & \multirowcell{3}{MAE\\$(\downarrow)$} & \multirowcell{3}{SSIM\\$(\uparrow)$} & \multirowcell{3}{PSNR\\$(\uparrow)$} \\ 
		& & & & & \\
		& & & & & \\ \midrule
		& raw & 0 & 0.092 & 0.778 & 18.645 \\ \midrule
		\multirowcell{2}{\rotatebox[origin=c]{90}{NNA}} & random & 2 & 0.111 & 0.740 & 17.386 \\ 
		& sub-hybrid & 2 & 0.181 & 0.519 & 13.434 \\ \midrule
		\multirowcell{3}{\rotatebox[origin=c]{90}{DIA}} & \multirowcell{3}{sub-hybrid} & 2 & 0.150 & 0.653 & 14.959 \\
		& & 4 & 0.160 & 0.611 & 14.471 \\
		& & 6 & 0.166 & 0.585 & 14.154 \\
		\bottomrule
	\end{tabular}
	\caption{{\bf Image reconstruction -- SIFT statistics.} We report quality metrics between reconstructed and original images for SIFT descriptors.}
	\label{tab:image_rec_sift_stat}
	\vspace{-5pt}
\end{table}
\setlength{\tabcolsep}{\tabcolsepdefault}

\customparagraph{Oracle Attack.}
In this section, we also provide the adversary with a fictional oracle that, given a list of possible attack descriptors for a private feature, returns the closest one to the original descriptor.
We propose the following attack methodology: for each private representation $\mathcal{D}$ associated to a descriptor $d$, the database $V$ of \numprint{128000} real-world descriptors is used to retrieve the $K$ closest elements to the subspace $\tilde{d}_1, \dots, \tilde{d}_K$.
Next, these attack descriptors are provided to the oracle, which returns the closest one to the original descriptor $d$, \ie,  $j = \arg\min_{i \in \{1, \dots, K\}} \lVert \tilde{d}_i - d \rVert$.
The descriptor $\tilde{d}_j$ is then used as an approximation to the original descriptor.
We also consider a version where the reconstructed descriptor is obtained by orthogonal projection of $\tilde{d}_j$ to the subspace $\mathcal{D}$ (denoted by \textit{proj.}).
Finally, a feature inversion network can be used to reconstruct the original image from the approximated descriptors.
In practice, the attacker does not have access to the original descriptor $d$, so implementing an oracle would be extremely challenging.

\begin{figure}
	\centering
	\begin{minipage}{.45\columnwidth}
		\centering
		 \includegraphics[width=\textwidth]{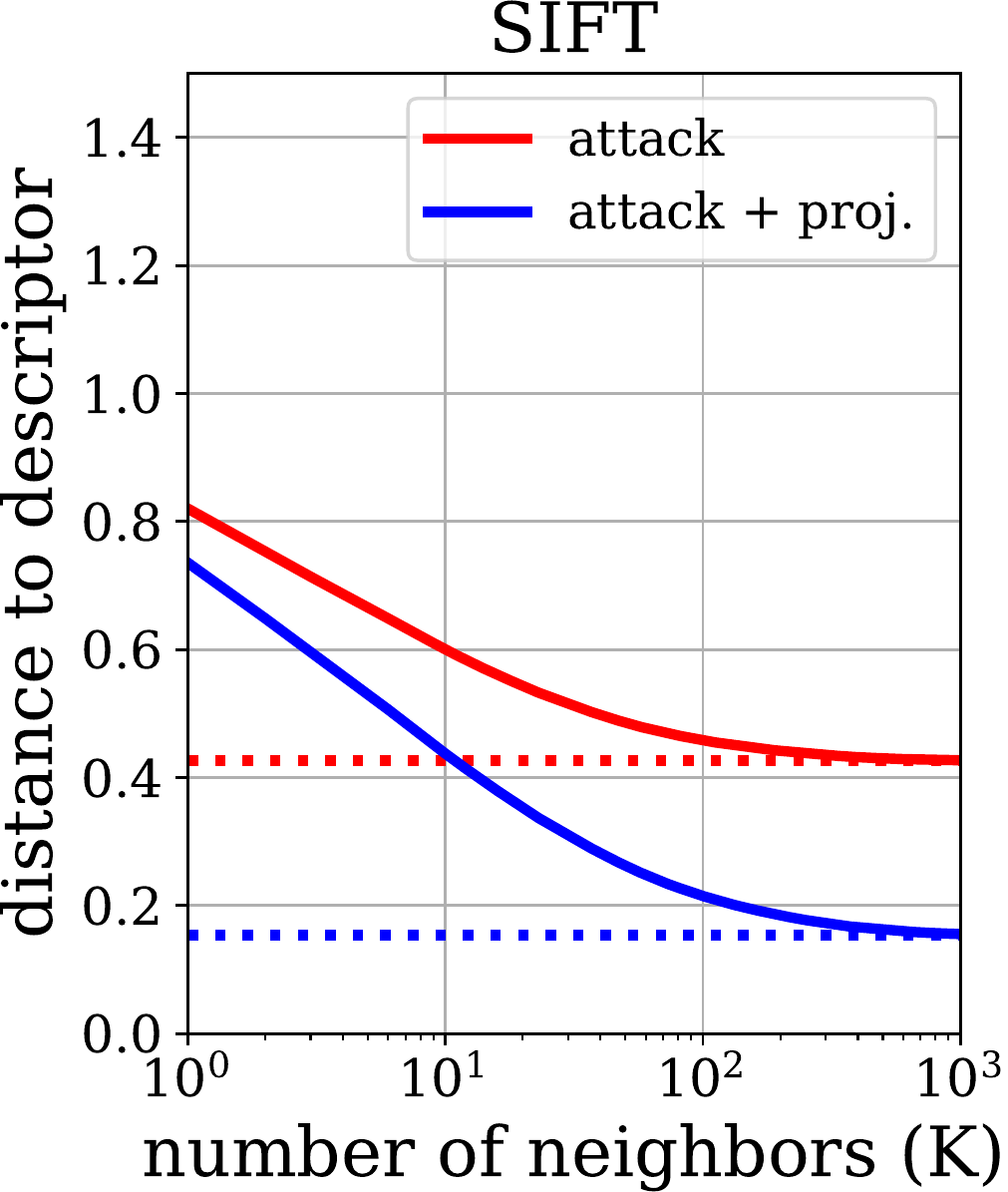}
	\end{minipage}
	~
	\begin{minipage}{.45\columnwidth}
		\centering
		\includegraphics[width=\textwidth]{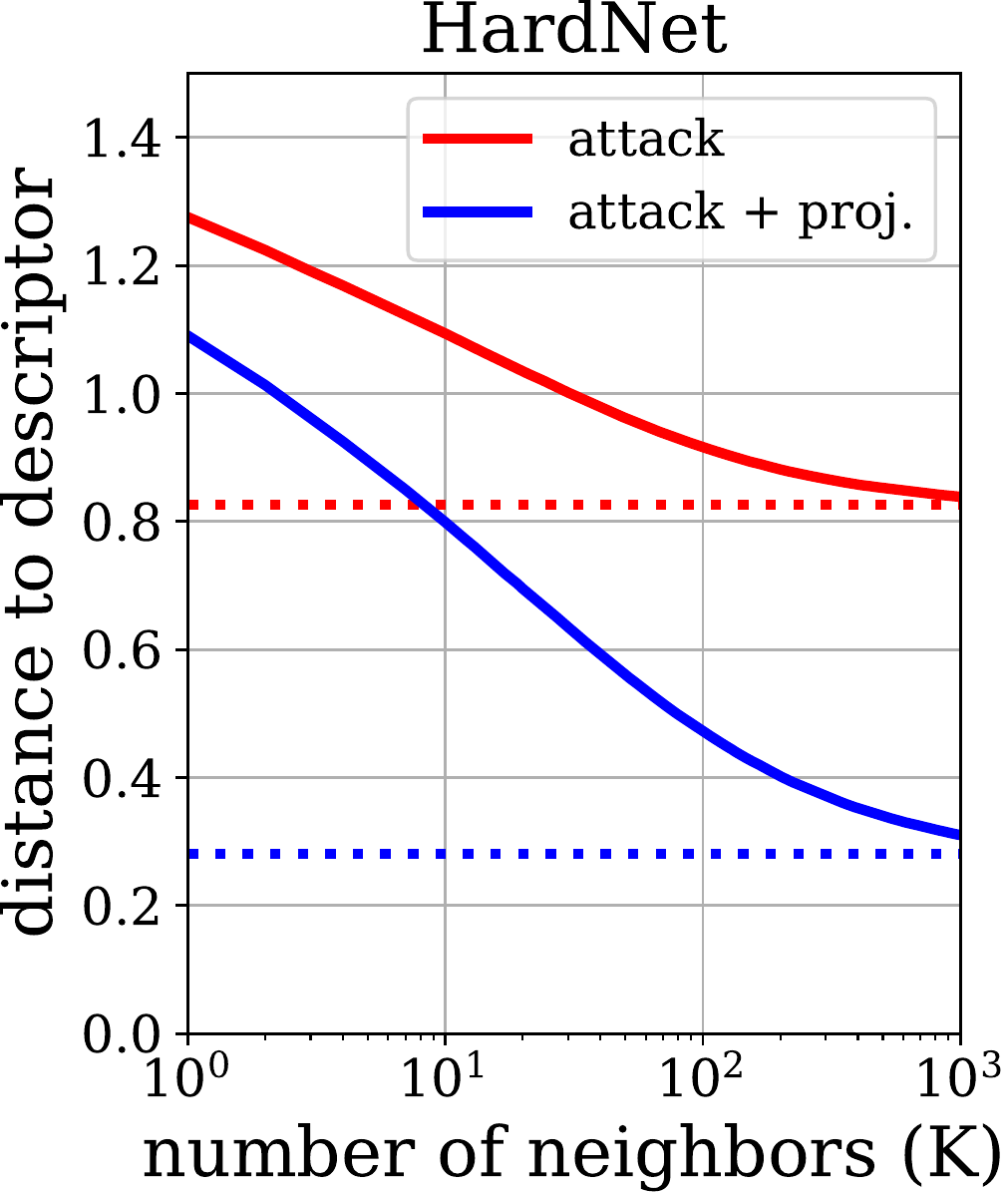}
	\end{minipage}
	\vspace{-5pt}
	\caption{{\bf Image reconstruction -- quantitative.} We plot the average distance of the reconstructed descriptors to the original one for different values of $K$, the number of nearest neighbors considered during the attack -- descriptors were lifted using sub-hybrid lifting to planes. The dotted lines are the limits of their solid conterparts.}
	\label{fig:image_rec_oracle_quantitative}
	\vspace{-5pt}
\end{figure}

\begin{figure*}[p]
	\centering
	\includegraphics[width=.9\textwidth]{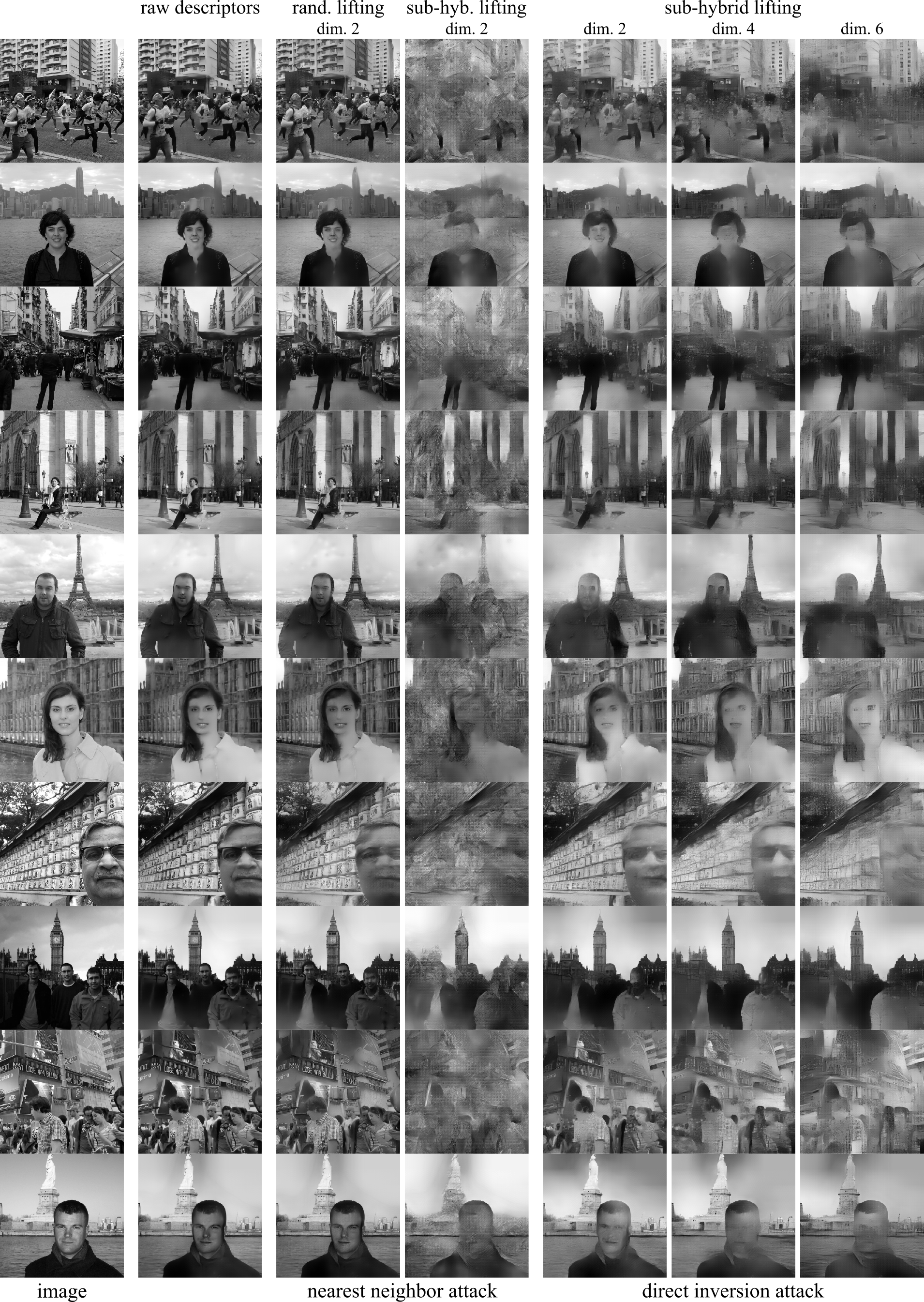}
	\vspace{-10pt}
	\caption{{\bf Image reconstruction -- SIFT.} We show qualitative examples: first original image, then reconstructions from the raw descriptors and using the proposed privacy attacks on different lifting methods and dimensions.}
	\label{fig:image_rec_qualitative_sift}
\end{figure*}

\begin{figure*}[p]
	\centering
	\includegraphics[width=.9\textwidth]{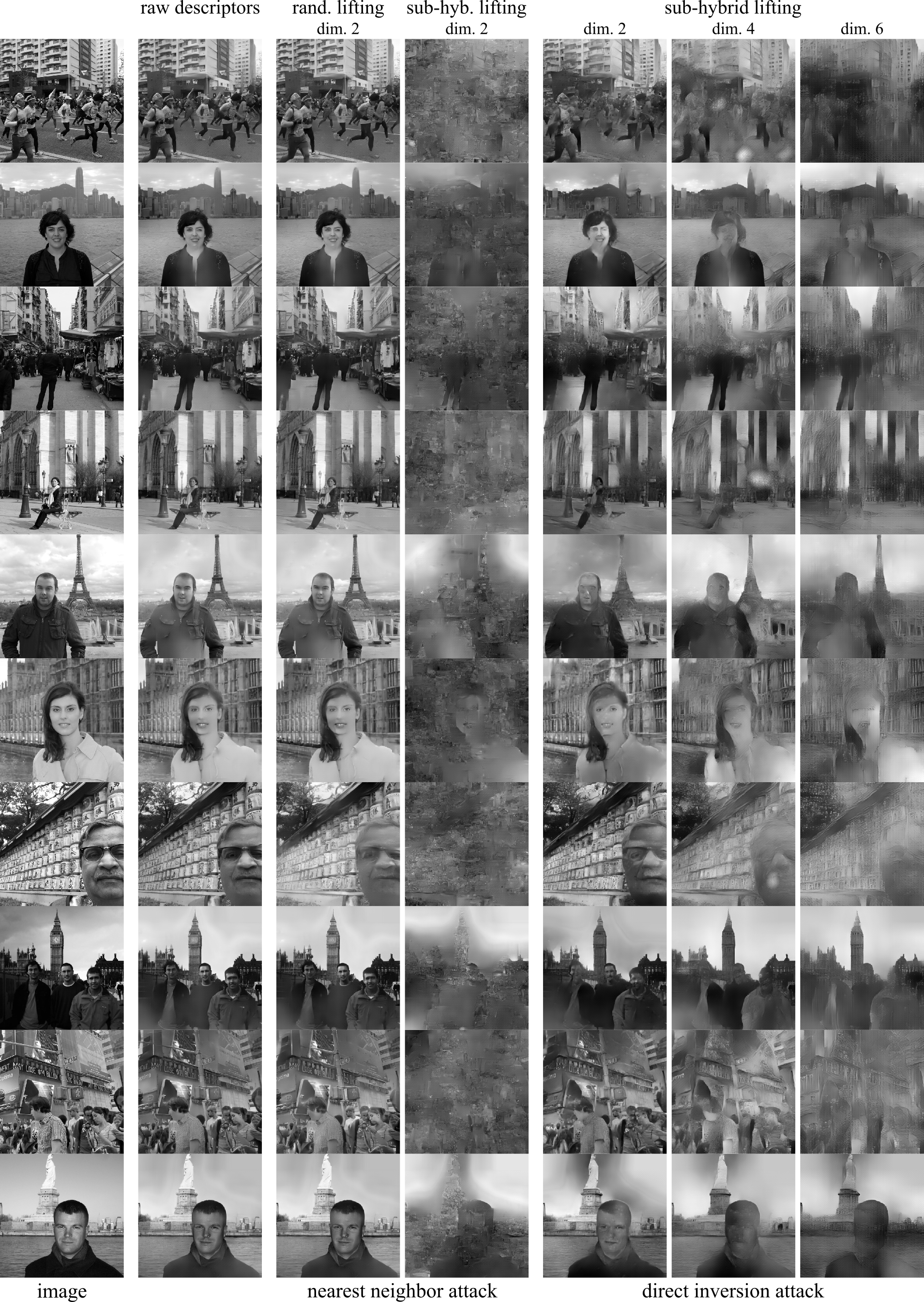}
	\vspace{-10pt}
	\caption{{\bf Image reconstruction -- HardNet.} We show qualitative examples: first original image, then reconstructions from the raw descriptors and using the proposed privacy attacks on different lifting methods and dimensions.}
	\label{fig:image_rec_qualitative_hardnet}
\end{figure*}

\begin{figure*}[p]
	\centering
	\includegraphics[width=.9\textwidth]{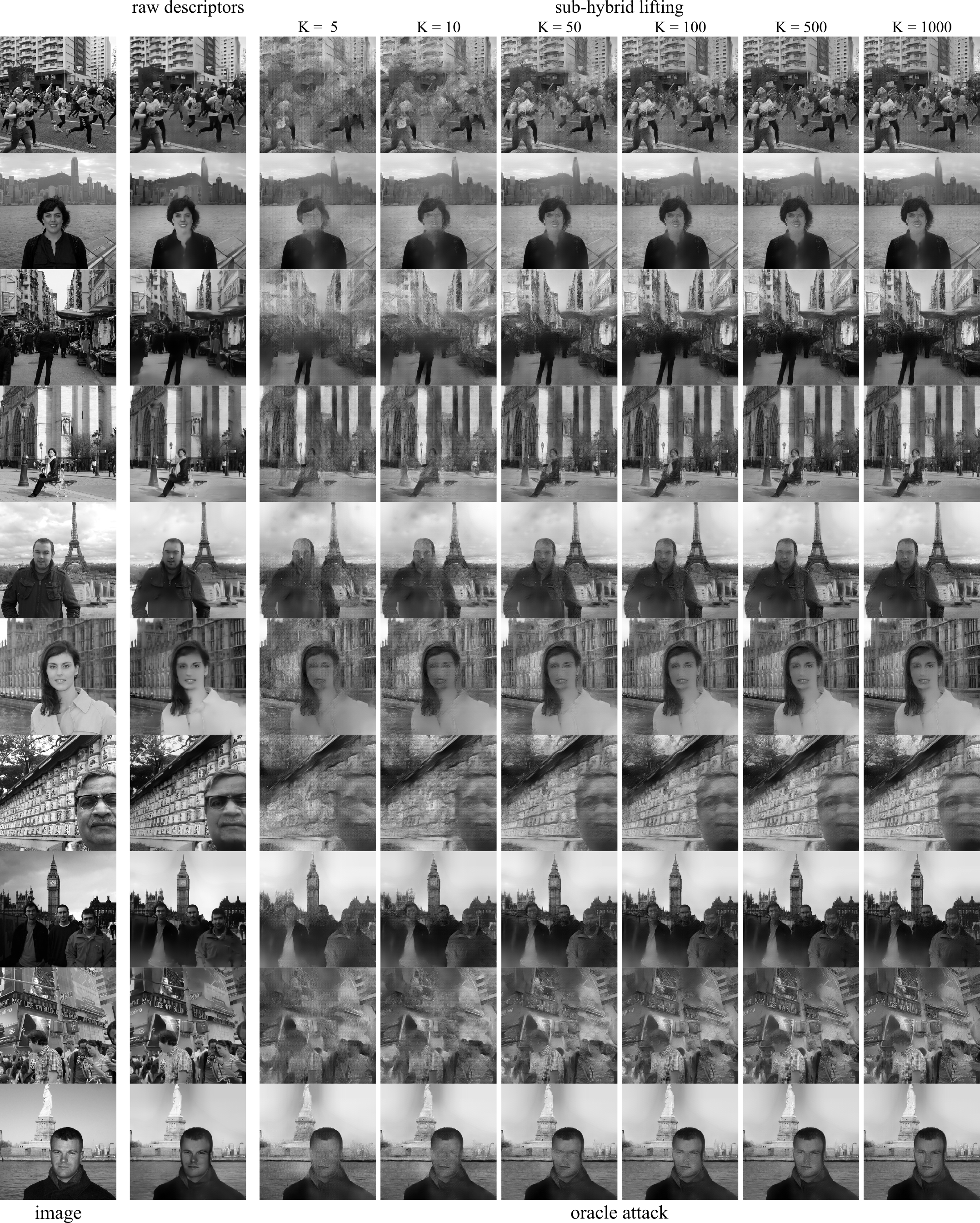}
	\vspace{-10pt}
	\caption{{\bf Image reconstruction (oracle) -- SIFT.} We show qualitative examples: first original image, then reconstructions from the raw descriptors and using the oracle privacy attack for different values of $K$. Descriptors are lifted to planes ($m = 2$).}
	\label{fig:image_rec_qualitative_sift_oracle}
\end{figure*}

\begin{figure*}[p]
	\centering
	\includegraphics[width=.9\textwidth]{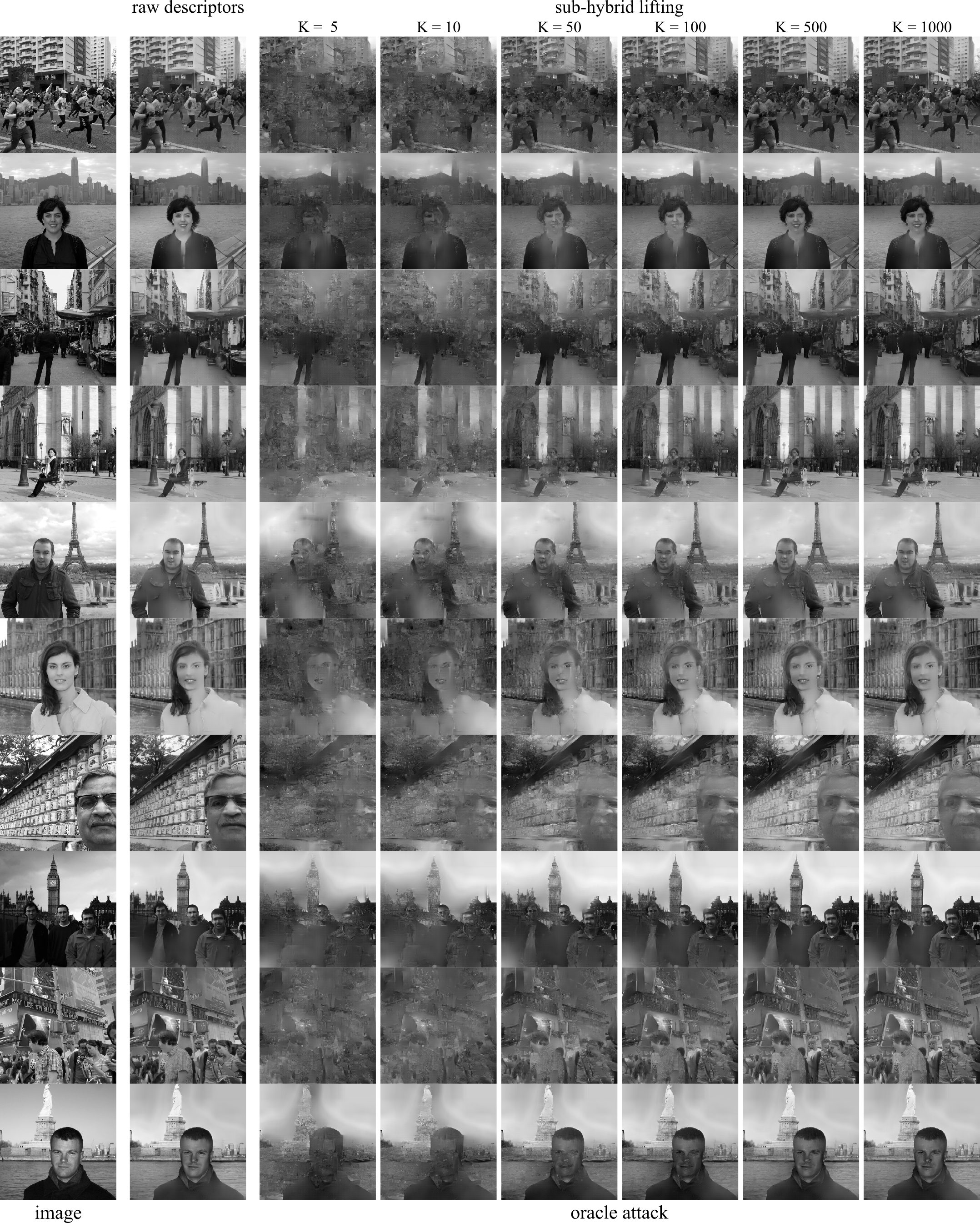}
	\vspace{-10pt}
	\caption{{\bf Image reconstruction (oracle) -- HardNet.} We show qualitative examples: first original image, then reconstructions from the raw descriptors and using the oracle privacy attack for different values of $K$. Descriptors are lifted to planes ($m = 2$).}
	\label{fig:image_rec_qualitative_hardnet_oracle}
\end{figure*}

Figure~\ref{fig:image_rec_oracle_quantitative} shows quantitative results of the oracle attack on the $10$ Flickr holiday images totaling around \numprint{40000} features with SIFT and HardNet descriptors.
We plot the average distance between the original and the reconstructed descriptor as a function of the number of neighbors $K$.
For this experiment, we used sub-hybrid lifting to planes ($m = 2$).
The projected version is always closer, but it is not necessarily on the unit hyper-sphere.
The dotted lines represent the asymptotic values of each respective curve, \ie, the value for $K=$ \numprint{128000}.
A first important observation is that, despite only using one adversarial sample during the subspace construction, there is a significant number of confounding real-world descriptors in the neighborhood of the subspace.
Note that SIFT descriptors only take positive values (\ie, in $\mathbb{R}_+^{128}$), which explains the smaller distance between reconstructed and original when compared to HardNet descriptors taking values in $\mathbb{R}^{128}$.
We also show qualitative examples in Figures~\ref{fig:image_rec_qualitative_sift_oracle} and~\ref{fig:image_rec_qualitative_hardnet_oracle}.
Even for large numbers of neighbors and access to an imaginary oracle, the reconstructed image remains far from the original.

\FloatBarrier
\newpage
{\small
\bibliographystyle{unsrt}
\bibliography{shortstrings,references}

\begin{thebibliography}{10}

\bibitem{Lowe2004Distinctive}
David~G. Lowe.
\newblock Distinctive image features from scale-invariant keypoints.
\newblock {\em IJCV}, 2004.

\bibitem{Agarwal2011Building}
Sameer Agarwal, Yasutaka Furukawa, Noah Snavely, Ian Simon, Brian Curless,
  Steven~M. Seitz, and Richard Szeliski.
\newblock {Building Rome in a Day}.
\newblock {\em Communications of the ACM}, 2011.

\bibitem{Sivic2003Video}
Josef Sivic and Andrew Zisserman.
\newblock {Video Google: A text retrieval approach to object matching in
  videos}.
\newblock In {\em Proc. ICCV}, 2003.

\bibitem{Turk1991Eigenfaces}
Matthew Turk and Alex Pentland.
\newblock {Eigenfaces for Recognition}.
\newblock {\em Journal of Cognitive Neuroscience}, 1991.

\bibitem{Dalal2005Histograms}
Navneet Dalal and Bill Triggs.
\newblock Histograms of oriented gradients for human detection.
\newblock In {\em Proc. CVPR}, 2005.

\bibitem{Boureau2010Learning}
Y-Lan Boureau, Francis Bach, Yann Le~Cun, and Jean Ponce.
\newblock Learning mid-level features for recognition.
\newblock In {\em Proc. CVPR}, 2010.

\bibitem{Sharif2014CNN}
Ali Sharif~Razavian, Hossein Azizpour, Josephine Sullivan, and Stefan Carlsson.
\newblock {CNN Features off-the-shelf: an Astounding Baseline for Recognition}.
\newblock In {\em Proc. CVPR Workshops}, 2014.

\bibitem{Zeiler2014Visualizing}
Matthew~D. Zeiler and Rob Fergus.
\newblock {Visualizing and Understanding Convolutional Networks}.
\newblock In {\em Proc. ECCV}, 2014.

\bibitem{Weinzaepfel2011Reconstructing}
Philippe Weinzaepfel, Herv{\'e} J{\'e}gou, and Patrick P{\'e}rez.
\newblock Reconstructing an image from its local descriptors.
\newblock In {\em Proc. CVPR}, 2011.

\bibitem{Dosovitskiy2016Inverting}
Alexey Dosovitskiy and Thomas Brox.
\newblock Inverting visual representations with convolutional networks.
\newblock In {\em Proc. CVPR}, 2016.

\bibitem{Zhmoginov2016Inverting}
Andrey Zhmoginov and Mark Sandler.
\newblock Inverting face embeddings with convolutional neural networks.
\newblock {\em arXiv}, 2016.

\bibitem{Mai2018Reconstruction}
Guangcan Mai, Kai Cao, Pong~C. Yuen, and Anil~K. Jain.
\newblock On the reconstruction of face images from deep face templates.
\newblock {\em IEEE PAMI}, 2018.

\bibitem{Paillier1999Public}
{Paillier, Pascal}.
\newblock {Public-key Cryptosystems Based on Composite Degree Residuosity
  Classes}.
\newblock In {\em Proc. EuroCrypt}, 1999.

\bibitem{Dwork2006Calibrating}
Cynthia Dwork, Frank McSherry, Kobbi Nissim, and Adam Smith.
\newblock Calibrating noise to sensitivity in private data analysis.
\newblock In {\em Proc. TCC}, 2006.

\bibitem{Kairouz2019Advances}
Peter Kairouz, H.~Brendan McMahan, Brendan Avent, Aur{\'e}lien Bellet, Mehdi
  Bennis, Arjun~Nitin Bhagoji, Keith Bonawitz, Zachary Charles, Graham Cormode,
  Rachel Cummings, et~al.
\newblock {Advances and Open Problems in Federated Learning}.
\newblock {\em arXiv}, 2019.

\bibitem{Speciale2019a}
Pablo Speciale, Johannes~L. Schonberger, Sing~Bing Kang, Sudipta~N. Sinha, and
  Marc Pollefeys.
\newblock {Privacy Preserving Image-Based Localization}.
\newblock In {\em Proc. CVPR}, 2019.

\bibitem{Speciale2019b}
Pablo Speciale, Johannes~L. Schonberger, Sudipta~N. Sinha, and Marc Pollefeys.
\newblock {Privacy Preserving Image Queries for Camera Localization}.
\newblock In {\em Proc. ICCV}, 2019.

\bibitem{Calonder2011BRIEF}
Michael Calonder, Vincent Lepetit, Mustafa Ozuysal, Tomasz Trzcinski, Christoph
  Strecha, and Pascal Fua.
\newblock {BRIEF: Computing a local binary descriptor very fast}.
\newblock {\em IEEE PAMI}, 2011.

\bibitem{Balntas2016Learning}
Vassileios Balntas, Edgar Riba, Daniel Ponsa, and Krystian Mikolajczyk.
\newblock Learning local feature descriptors with triplets and shallow
  convolutional neural networks.
\newblock In {\em Proc. BMVC.}, 2016.

\bibitem{He2018Local}
Kun He, Yan Lu, and Stan Sclaroff.
\newblock {Local Descriptors Optimized for Average Precision}.
\newblock In {\em Proc. CVPR}, 2018.

\bibitem{Mishchuk2017Working}
Anastasiya Mishchuk, Dmytro Mishkin, Filip Radenovic, and Jiri Matas.
\newblock Working hard to know your neighbor's margins: Local descriptor
  learning loss.
\newblock In {\em Advances in NeurIPS}, 2017.

\bibitem{Jegou2010Aggregating}
Herv{\'e} J{\'e}gou, Matthijs Douze, Cordelia Schmid, and Patrick P{\'e}rez.
\newblock Aggregating local descriptors into a compact image representation.
\newblock In {\em Proc. CVPR}, 2010.

\bibitem{schoenberger2015detail}
Johannes~Lutz Sch\"{o}nberger, Filip Radenovi\'{c}, Ondrej Chum, and
  Jan-Michael Frahm.
\newblock {From Single Image Query to Detailed 3D Reconstruction}.
\newblock In {\em Conference on Computer Vision and Pattern Recognition
  (CVPR)}, 2015.

\bibitem{Zhang2016Joint}
Kaipeng Zhang, Zhanpeng Zhang, Zhifeng Li, and Yu~Qiao.
\newblock {Joint Face Detection and Alignment Using Multi-task Cascaded
  Convolutional Networks}.
\newblock {\em IEEE Signal Processing Letters}, 2016.

\bibitem{Taigman2014DeepFace}
Yaniv Taigman, Ming Yang, Marc'Aurelio Ranzato, and Lior Wolf.
\newblock {DeepFace: Closing the gap to human-level performance in face
  verification}.
\newblock In {\em Proc. CVPR}, 2014.

\bibitem{Liu2017SphereFace}
Weiyang Liu, Yandong Wen, Zhiding Yu, Ming Li, Bhiksha Raj, and Le~Song.
\newblock {SphereFace: Deep Hypersphere Embedding for Face Recognition}.
\newblock In {\em Proc. CVPR}, 2017.

\bibitem{Deng2018ArcFace}
Jiankang Deng, Jia Guo, Xue Niannan, and Stefanos Zafeiriou.
\newblock {ArcFace: Additive Angular Margin Loss for Deep Face Recognition}.
\newblock In {\em Proc. CVPR}, 2019.

\bibitem{Wang2014Affine}
Zhenhua Wang, Bin Fan, and Fuchao Wu.
\newblock {Affine Subspace Representation for Feature Description}.
\newblock In {\em Proc. ECCV}, 2014.

\bibitem{Dosovitskiy2016Generating}
Alexey Dosovitskiy and Thomas Brox.
\newblock Generating images with perceptual similarity metrics based on deep
  networks.
\newblock In {\em Advances in NeurIPS}, 2016.

\bibitem{Pittaluga2019}
Francesco Pittaluga, Sanjeev~J Koppal, Sing Bing~Kang, and Sudipta~N Sinha.
\newblock {Revealing Scenes by Inverting Structure From Motion
  Reconstructions}.
\newblock In {\em Proc. CVPR}, 2019.

\bibitem{Dalenius1977Towards}
Tore Dalenius.
\newblock Towards a methodology for statistical disclosure control.
\newblock {\em Statistik Tidskrift}, 1977.

\bibitem{Dwork2008Differential}
Cynthia Dwork.
\newblock {Differential Privacy: A Survey of Results}.
\newblock In {\em Proc. TAMC}, 2008.

\bibitem{McMahan2016Communication}
H.~Brendan McMahan, Eider Moore, Daniel Ramage, Seth Hampson, and Blaise
  Ag\"uera~y Arcas.
\newblock {Communication-Efficient Learning of Deep Networks from Decentralized
  Data}.
\newblock {\em arXiv}, 2016.

\bibitem{Qin2014Towards}
Zhan Qin, Jingbo Yan, Kui Ren, Chang~Wen Chen, and Cong Wang.
\newblock Towards efficient privacy-preserving image feature extraction in
  cloud computing.
\newblock In {\em Proc. ACMM}, 2014.

\bibitem{Hsu2012Image}
Chao-Yung Hsu, Chun-Shien Lu, and Soo-Chang Pei.
\newblock {Image feature extraction in encrypted domain with privacy-preserving
  SIFT}.
\newblock {\em IEEE Transactions on Image Processing}, 2012.

\bibitem{Jiang2017Secure}
Linzhi Jiang, Chunxiang Xu, Xiaofang Wang, Bo~Luo, and Huaqun Wang.
\newblock {Secure outsourcing SIFT: Efficient and privacy-preserving image
  feature extraction in the encrypted domain}.
\newblock {\em IEEE Transactions on Dependable and Secure Computing}, 2017.

\bibitem{Kim2020Efficient}
Taeyun Kim, Yongwoo Oh, and Hyoungshick Kim.
\newblock {Efficient Privacy-Preserving Fingerprint-Based Authentication System
  Using Fully Homomorphic Encryption}.
\newblock {\em Security and Communication Networks}, 2020.

\bibitem{boddeti2018secure}
Vishnu~Naresh Boddeti.
\newblock Secure face matching using fully homomorphic encryption.
\newblock In {\em International Conference on Biometrics Theory, Applications
  and Systems}, 2018.

\bibitem{engelsma2020hers}
Joshua~J Engelsma, Anil~K Jain, and Vishnu~Naresh Boddeti.
\newblock Hers: Homomorphically encrypted representation search.
\newblock {\em arXiv}, 2020.

\bibitem{geppert2020privacy}
Marcel Geppert, Viktor Larsson, Pablo Speciale, Johannes~Lutz Sch\"{o}nberger,
  and Marc Pollefeys.
\newblock {Privacy Preserving Structure-from-Motion}.
\newblock In {\em Proc. ECCV}, 2020.

\bibitem{shibuya2020privacy}
Mikiya Shibuya, Shinya Sumikura, and Ken Sakurada.
\newblock Privacy preserving visual {SLAM}.
\newblock In {\em Proc. ECCV}, 2020.

\bibitem{Balntas2017HPatches}
Vassileios Balntas, Karel Lenc, Andrea Vedaldi, and Krystian Mikolajczyk.
\newblock {HPatches}: A benchmark and evaluation of handcrafted and learned
  local descriptors.
\newblock In {\em Proc. CVPR}, 2017.

\bibitem{Zhou2017Places}
Bolei Zhou, Agata Lapedriza, Aditya Khosla, Aude Oliva, and Antonio Torralba.
\newblock Places: A 10 million image database for scene recognition.
\newblock {\em IEEE PAMI}, 2017.

\bibitem{Bishop2006Pattern}
Christopher~M. Bishop.
\newblock {\em {Pattern Recognition and Machine Learning}}.
\newblock Springer, 2006.

\bibitem{Schonberger2017Comparative}
Johannes~L. Sch\"onberger, Hans Hardmeier, Torsten Sattler, and Marc Pollefeys.
\newblock Comparative evaluation of hand-crafted and learned local features.
\newblock In {\em Proc. CVPR}, 2017.

\bibitem{Dusmanu2019D2}
Mihai Dusmanu, Ignacio Rocco, Tomas Pajdla, Marc Pollefeys, Josef Sivic,
  Akihiko Torii, and Torsten Sattler.
\newblock {D2-Net: A Trainable CNN for Joint Detection and Description of Local
  Features}.
\newblock In {\em Proc. CVPR}, 2019.

\bibitem{Schoenberger2016Structure}
Johannes~Lutz Sch\"{o}nberger and Jan-Michael Frahm.
\newblock {Structure-from-Motion Revisited}.
\newblock In {\em Proc. CVPR}, 2016.

\bibitem{Arandjelovic2016NetVLAD}
Relja Arandjelovic, Petr Gronat, Akihiko Torii, Tomas Pajdla, and Josef Sivic.
\newblock {NetVLAD}: {CNN} architecture for weakly supervised place
  recognition.
\newblock In {\em Proc. CVPR}, 2016.

\bibitem{schoenberger2016vote}
Johannes~Lutz Sch\"{o}nberger, True Price, Torsten Sattler, Jan-Michael Frahm,
  and Marc Pollefeys.
\newblock {A Vote-and-Verify Strategy for Fast Spatial Verification in Image
  Retrieval}.
\newblock In {\em Asian Conference on Computer Vision (ACCV)}, 2016.

\bibitem{Sattler2017Benchmarking}
Torsten Sattler, Will Maddern, Carl Toft, Akihiko Torii, Lars Hammarstrand,
  Erik Stenborg, Daniel Safari, Masatoshi Okutomi, Marc Pollefeys, Josef Sivic,
  Fredrik Kahl, and Tomas Pajdla.
\newblock Benchmarking {6DoF} outdoor visual localization in changing
  conditions.
\newblock In {\em Proc. CVPR}, 2018.

\bibitem{Google2019Google}
Tilman Reinhardt.
\newblock {Google Visual Positioning Service}.
\newblock
  \url{https://ai.googleblog.com/2019/02/using-global-localization-to-improve.html},
  2019.

\bibitem{Microsoft2019Announcing}
Neena Kamath.
\newblock {Announcing Azure Spatial Anchors for collaborative, cross-platform
  mixed reality apps}.
\newblock
  \url{https://azure.microsoft.com/en-us/blog/announcing-azure-spatial-anchors-for-collaborative-cross-platform-mixed-reality-apps/},
  2019.

\bibitem{VisualLocalization}
{Long-Term Visual Localization Benchmark}.
\newblock \url{https://www.visuallocalization.net}.

\bibitem{Li2018MegaDepth}
Zhengqi Li and Noah Snavely.
\newblock {MegaDepth}: Learning single-view depth prediction from internet
  photos.
\newblock In {\em Proc. CVPR}, 2018.

\bibitem{He2016Deep}
Kaiming He, Xiangyu Zhang, Shaoqing Ren, and Jian Sun.
\newblock Deep residual learning for image recognition.
\newblock In {\em Proc. CVPR}, 2016.

\bibitem{Guo2016MS}
Yandong Guo, Lei Zhang, Yuxiao Hu, Xiaodong He, and Jianfeng Gao.
\newblock {MS-Celeb-1M: Challenge of Recognizing One Million Celebrities in the
  Real World}.
\newblock {\em Journal of Electronic Imaging}, 2016.

\bibitem{Huang2007Labeled}
Gary~B. Huang, Marwan Mattar, Tamara Berg, and Eric Learned-Miller.
\newblock Labeled faces in the wild: A database for studying face recognition
  in unconstrained environments.
\newblock Technical report, University of Massachusetts, Amherst, 2007.

\bibitem{Sengupta2016Frontal}
Soumyadip Sengupta, Jun-Cheng Chen, Carlos Castillo, Vishal~M. Patel, Rama
  Chellappa, and David~W. Jacobs.
\newblock Frontal to profile face verification in the wild.
\newblock In {\em Proc. WACV}, 2016.

\bibitem{Moschoglou2017AgeDB}
Stylianos Moschoglou, Athanasios Papaioannou, Christos Sagonas, Jiankang Deng,
  Irene Kotsia, and Stefanos Zafeiriou.
\newblock {AgeDB: The First Manually Collected, In-The-Wild Age Database}.
\newblock In {\em Proc. CVPR Workshops}, 2017.

\bibitem{WindowsHello}
{Windows Hello}.
\newblock
  \url{https://blogs.windows.com/windowsexperience/2015/07/25/say-hello-to-windows-hello-on-windows-10/}.

\bibitem{FaceID}
{Face ID}.
\newblock \url{https://support.apple.com/en-us/HT208108}.

\bibitem{Karkkainen2019Fair}
Kimmo K\"arkk\"ainen and Jungseock Joo.
\newblock {FairFace: Face Attribute Dataset for Balanced Race, Gender, and
  Age}.
\newblock {\em arXiv}, 2019.

\bibitem{Basri2007Approximate}
Ronen Basri, Tal Hassner, and Lihi Zelnik-Manor.
\newblock {Approximate Nearest Subspace Search with Applications to Pattern
  Recognition}.
\newblock In {\em Proc. CVPR}, 2007.

\end{thebibliography}
}
	
\end{document}